%% file: main.tex
\pdfoutput=1
\documentclass[11pt]{article}

\usepackage{acl}

\usepackage{times}
\usepackage{latexsym}
\usepackage[T1]{fontenc}
\usepackage[utf8]{inputenc}
\usepackage{microtype}
\usepackage{amsmath,amssymb,amsthm}
\newtheorem{proposition}{Proposition}
\usepackage{graphicx}
\usepackage{booktabs}
\usepackage{multirow}
\usepackage{xcolor}
\usepackage{hyperref}
\usepackage{tikz}
\usetikzlibrary{positioning, arrows.meta, shapes.geometric, calc, decorations.pathreplacing, fit, backgrounds}

\newcommand{\Yhat}{\hat{Y}}
\newcommand{\Ytilde}{\widetilde{Y}}
\newcommand{\Ahat}{\hat{A}}
\newcommand{\Atilde}{\widetilde{A}}
\newcommand{\pitilde}{\widetilde{\pi}}
\newcommand{\thetahat}{\hat{\theta}}
\newcommand{\thetaref}{\theta^*}
\newcommand{\That}{\hat{T}}
\newcommand{\Tref}{T^*}
\newcommand{\deltahat}{\hat{\delta}}
\newcommand{\RMSE}{\mathrm{RMSE}}
\newcommand{\sRMSE}{\mathrm{sRMSE}}
\newcommand{\E}{\mathbb{E}}
\newcommand{\R}{\mathbb{R}}
\newcommand{\1}{\mathbf{1}}
\newcommand{\clip}{\operatorname{clip}}

\title{\textsc{Lucid}: Statistically Grounded Self-Correction \\ for LLM Reasoning Evaluation}

\author{%
  Satiyabooshan Murugaboopathy$^{1,2,*}$ \quad 
  Adel Daoud$^{1,2,3}$ \quad 
  Connor T.\ Jerzak$^{1,4}$ \\[6pt]
  \small $^1$AI and Global Development Lab \\
  \small $^2$Institute for Analytical Sociology, Link\"oping University, Link\"oping, Sweden \\
  \small $^3$NLP@DSAI, Chalmers University of Technology, Gothenburg, Sweden \\
  \small $^4$Department of Government, The University of Texas at Austin, Austin, TX, USA \\
  \small $^*$Corresponding author: \texttt{satiyabooshan.murugaboopathy@liu.se}
}

\begin{document}
\AddToHookNext{shipout/foreground}{%
  \setlength{\unitlength}{1bp}%
  \put(297.638,-805.04){\makebox[0pt][c]{\fontsize{9}{11}\selectfont\itshape
    Findings of the Association for Computational Linguistics: EACL 2027}}%
  \put(297.638,-816.00){\makebox[0pt][c]{\fontsize{9}{11}\selectfont
    March 9-14, 2027 \textcopyright 2027 Association for Computational Linguistics}}%
}
\maketitle

\input{sections/00_abstract}

\input{figures/fig1_showcase}

\input{sections/01_introduction}
\input{sections/02_related_work}

\input{sections/03_methodology}
\input{sections/04_experiments}
\input{sections/05_results}
\input{sections/06_conclusion}
\input{sections/07_limitations}
\input{sections/08_ethical_considerations}

\bibliography{references}

\appendix
\input{sections/09_appendix_background}
\input{sections/10_appendix_features}
\input{sections/11_appendix_proof_double_robustness}
\input{sections/12_appendix_iterative_dsl}
\input{sections/13_appendix_adaptive}
\input{sections/14_appendix_rag}
\input{sections/15_appendix_sensitivity}
\input{sections/16_appendix_bias_analysis}
\input{sections/17_appendix_transition_matrix}
\input{sections/18_appendix_self_correction}
\input{sections/19_appendix_additional_figures}

\end{document}

%% file: sections/00_abstract.tex
\begin{abstract}
Cheap language-model raters grade chain-of-thought reasoning at scale, yet systematic step-rating errors can skew reasoning audits and solution rankings. We introduce \textsc{Lucid} (\textbf{L}oop for \textbf{U}nbiased \textbf{C}alibration via \textbf{I}nference-time \textbf{D}iagnosis), which pairs prompt self-correction with Design-based Supervised Learning (DSL). Without retraining or weight access, \textsc{Lucid} calibrates binary step ratings by aligning predicted and expert soundness rates. From a small expert-labeled set, \textsc{Lucid} summarizes rater bias in plain text and prompts the rater to re-grade unseen steps. For example, on PRM800K, GPT-5-mini grades too strictly, under-rating computation, verification, and late-trace steps most acutely. In one pass, \textsc{Lucid} shrinks GPT-5-mini's calibration gap from 7.8 to 0.9 percentage points in absolute predicted--expert rate difference. \textsc{Lucid} also halves false alarms (sound steps rated unsound) and lifts error-detection $F_{1,-}$ by 32\%, where $F_{1,-}$ measures $F_1$ on unsound steps. \textsc{Lucid} retains these gains with 1{,}000 diagnostic steps or overall-gap feedback alone. The loop also transfers across models and domains: it applies GPT-5-mini's bias profile to lift GLM-5.2's PRM800K accuracy, and it narrows calibration gaps for overly strict but capable raters across math, science, web, and logic benchmarks. Because prompt feedback shifts thresholds without improving class-balanced discrimination (balanced-class sound/unsound separation), \textsc{Lucid} filters re-grading with pre-/post-checks that skip already calibrated raters and roll back ineffective passes. Finally, we reuse that expert set with raw corpus ratings in DSL for calibration and regression debiasing, and apply DSL to pair-labeled steps for sound-step transitions. DSL's out-of-fold (cross-fitted) soundness probabilities lift best-of-$K$ selection to match expert labels, while its doubly-robust estimators beat expert-only regression fits at small budgets and cut sound-step transition error by nearly two-thirds. 
\end{abstract}

%% file: figures/fig1_showcase.tex
\definecolor{tuftegreen}{RGB}{32,122,72}
\definecolor{tuftecoral}{RGB}{198,64,58}
\definecolor{tufteslate}{RGB}{95,105,115}

\begin{figure*}[t]
\centering
\resizebox{\textwidth}{!}{%
\begin{tikzpicture}[
    x=1cm, y=1cm, >=Stealth,
    paneltitle/.style={font=\fontsize{9}{10.5}\selectfont\bfseries, text=black!88, anchor=west, inner sep=0pt},
    stepblock/.style={font=\fontsize{7}{8.8}\selectfont, text=black!86, align=left, anchor=north west, inner sep=0pt, text width=4.35cm},
    checkblock/.style={font=\fontsize{6.2}{7.6}\selectfont, text=black!48, align=left, anchor=north west, inner sep=0pt, text width=5.10cm},
    axislabel/.style={font=\fontsize{7}{8.2}\selectfont, text=black!78, inner sep=0pt},
    ticklabel/.style={font=\fontsize{6}{7.0}\selectfont, text=black!62, inner sep=0pt},
    majorlabel/.style={font=\fontsize{7}{8.0}\selectfont, inner sep=0pt},
    minorlabel/.style={font=\fontsize{6}{7.0}\selectfont, text=tufteslate!95!black, align=left, inner sep=0pt},
]

\node[paneltitle] at (0.00, 4.26) {(a) Inference-time calibration loop (\textsc{Lucid})};

\draw[line width=0.4pt, black!28] (0.00, 3.86) -- (4.35, 3.86);
\node[stepblock] at (0.00, 3.74) {%
    \textbf{1.\ Uncorrected rater} ($\Yhat_i^{(0)}$)\\[1.5pt]
    Under-rates computation \& late steps\\[1.5pt]
    {\color{black!62}\textit{e.g.,} correct arithmetic step $\mapsto$ {\color{tuftecoral}$\times$} (true {\color{tuftegreen}\checkmark})}%
};

\draw[line width=0.4pt, black!28] (5.15, 3.86) -- (9.50, 3.86);
\node[stepblock] at (5.15, 3.74) {%
    \textbf{2.\ Bias diagnosis + DSL}\\[1.5pt]
    Biases ($\deltahat_{\mathcal{D}}$, $\deltahat_{\mathcal{D},c}$, pos.\ $r$) \& DSL $\hat{\mu}_i, \Ytilde_i$%
};

\draw[line width=0.35pt, black!22, densely dashed] (5.15, 2.93) -- (9.50, 2.93);
\node[checkblock] at (5.15, 2.84) {%
    \textit{Pre-check} ($\mathcal{D}$): $|\deltahat_{\mathcal{D}}|{>}2\,\mathrm{SE},\,{>}2$pp?\\[1.2pt]
    {\color{tuftecoral!70}if no, fail} $\to$ skip (keep $\Yhat^{(0)}$)%
};

\draw[line width=0.4pt, black!28] (5.15, 1.84) -- (9.50, 1.84);
\node[stepblock] at (5.15, 1.72) {%
    \textbf{3.\ Calibration prompt}\\[1.5pt]
    Appends diagnosed bias as text%
};

\draw[line width=0.4pt, black!28] (0.00, 1.84) -- (4.35, 1.84);
\node[stepblock, text width=4.95cm] at (0.00, 1.72) {%
    \textbf{4.\ Content re-rating} ($\Yhat_i^{(1)}$)\\[1.5pt]
    Flips false alarm ({\color{tuftecoral}$\times$}${\to}${\color{tuftegreen}\checkmark}), catches true errors%
};

\draw[line width=0.35pt, black!22, densely dashed] (0.00, 0.77) -- (4.35, 0.77);
\node[checkblock] at (0.00, 0.67) {%
    \textit{Post-check} (fresh $\mathcal{V}$): $|\deltahat_{\mathcal{V}}^{(1)}|{<}|\deltahat_{\mathcal{V}}^{(0)}|$?\\[1.2pt]
    {\color{tuftegreen!72}if yes, pass} $\to$ keep $\Yhat^{(1)}$; {\color{tuftecoral!70}if no, fail} $\to$ revert $\Yhat^{(0)}$%
};

\draw[->, line width=0.55pt, black!68] (4.44, 3.45) -- (5.04, 3.45);
\draw[->, line width=0.42pt, black!36] (7.32, 3.16) -- (7.32, 2.96);
\draw[->, line width=0.42pt, black!36] (7.32, 2.26) -- (7.32, 1.89);
\node[font=\fontsize{6.2}{7.6}\selectfont, text=tuftegreen!72, anchor=west, inner sep=0pt] at (7.43, 2.08) {if yes, pass};
\draw[->, line width=0.55pt, black!68] (5.04, 1.54) -- (4.44, 1.54);
\draw[->, line width=0.42pt, black!36] (2.18, 1.07) -- (2.18, 0.81);
\draw[->, line width=0.45pt, black!38, dashed] (2.18, 1.94) -- (2.18, 2.74);

\node[paneltitle] at (10.30, 4.26) {(b) Calibration gap vs.\ unsound-step detection};

\draw[line width=0.45pt, black!60] (11.10, 0.72) -- (16.45, 0.72);
\draw[line width=0.45pt, black!60] (11.10, 0.72) -- (11.10, 3.52);

\foreach \g/\xlab in {0/0, 2/2, 4/4, 6/6, 8/8, 10/10} {
    \pgfmathsetmacro{\xp}{11.10 + 0.535*\g}
    \draw[line width=0.4pt, black!55] (\xp, 0.72) -- (\xp, 0.64);
    \node[ticklabel, anchor=north] at (\xp, 0.59) {\xlab};
}
\node[axislabel] at (13.78, 0.16) {Calibration gap $|\deltahat_{\mathcal{E}}|$ (pp, $\leftarrow$ better)};

\foreach \f/\ylab in {0.00/0, 0.05/.05, 0.10/.10, 0.15/.15, 0.20/.20} {
    \pgfmathsetmacro{\yp}{0.72 + 14.00*\f}
    \draw[line width=0.4pt, black!55] (11.10, \yp) -- (11.02, \yp);
    \node[ticklabel, anchor=east] at (10.97, \yp) {\ylab};
}
\node[axislabel, rotate=90] at (10.34, 2.12) {Error detection $F_{1,-}$ ($\rightarrow$ better)};

\draw[->, line width=0.45pt, tufteslate!75, densely dashed] (15.16, 2.74) -- (11.21, 1.91);
\draw[->, line width=0.45pt, tufteslate!75, densely dashed] (15.18, 2.70) -- (12.48, 1.83);
\draw[->, line width=0.45pt, tufteslate!75, densely dashed] (15.29, 2.66) -- (15.58, 0.83);

\draw[->, line width=1.05pt, tuftegreen] (15.16, 2.78) -- (11.71, 3.39);

\fill[tufteslate] (11.100, 1.882) circle (0.055);
\fill[tufteslate] (12.384, 1.784) circle (0.055);
\fill[tufteslate] (15.594, 0.720) circle (0.055);

\fill[white] (15.273, 2.764) circle (0.072);
\draw[tuftecoral, line width=0.9pt] (15.273, 2.764) circle (0.072);

\fill[tuftegreen] (11.582, 3.408) circle (0.078);

\node[majorlabel, text=tuftecoral!90!black, align=center, anchor=south] at (15.42, 2.88)
    {\textbf{Uncorrected}\\{\fontsize{6}{7.0}\selectfont\color{black!62}$7.8$\,pp, $0.146$}};

\node[majorlabel, text=tuftegreen!90!black, align=left, anchor=south west] at (11.74, 3.42)
    {\textbf{\textsc{Lucid} (1 round)}\\{\fontsize{6}{7.0}\selectfont $0.9$\,pp ($8.7\times$), $0.192$ ($+32\%$)}};

\draw[line width=0.3pt, black!35] (11.13, 1.96) -- (11.27, 2.34);
\node[minorlabel, anchor=south west] at (11.28, 2.37)
    {Global threshold};

\node[minorlabel, anchor=north west] at (12.35, 1.66)
    {Post-hoc type flips};

\node[minorlabel, anchor=south east] at (15.44, 0.80)
    {Platt / log.\ reg.};

\end{tikzpicture}%
}
\caption{\textbf{\textsc{Lucid} overview.}
\textbf{(a)}~From expert-labeled steps $\mathcal{D}$, \textsc{Lucid} diagnoses evaluator bias (overall $\deltahat_{\mathcal{D}}$, per-type $\deltahat_{\mathcal{D},c}$, position correlation $r$) and feeds it back for one re-rating pass ($\Yhat_i^{(0)}\!\to\!\Yhat_i^{(1)}$), with guards: a pre-check (cluster-robust $\mathrm{SE}$) skips calibrated raters, and a post-check on fresh expert-labeled validation set $\mathcal{V}$ reverts non-improving passes. DSL reuses $\mathcal{D}$ for cross-fitted calibration ($\hat{\mu}_i$) and doubly-robust regression debiasing ($\Ytilde_i$).
\textbf{(b)}~On PRM800K (eval set $\mathcal{E}$, $n_{\mathrm{eval}}{=}1{,}000$), non-LLM baselines (dashed: global rate shift, random per-type flips, Platt/logistic regression; Section~\ref{sec:iterative-prompt-results}, Table~\ref{tab:baselines}) either cut the calibration gap $|\deltahat_{\mathcal{E}}| = |\mathcal{E}|^{-1}\big|\sum_{i\in\mathcal{E}}(\Yhat_i - Y_i)\big|$ at the cost of $F_{1,-}$ ($F_1$ on unsound $Y_i{=}0$) or collapse to always-sound ($\Yhat_i$: rater; $Y_i$: expert). \textsc{Lucid} (solid green) re-grades step content to improve both ($8.7\times$ smaller gap, $+32\%$ $F_{1,-}$).}
\label{fig:showcase}
\end{figure*}
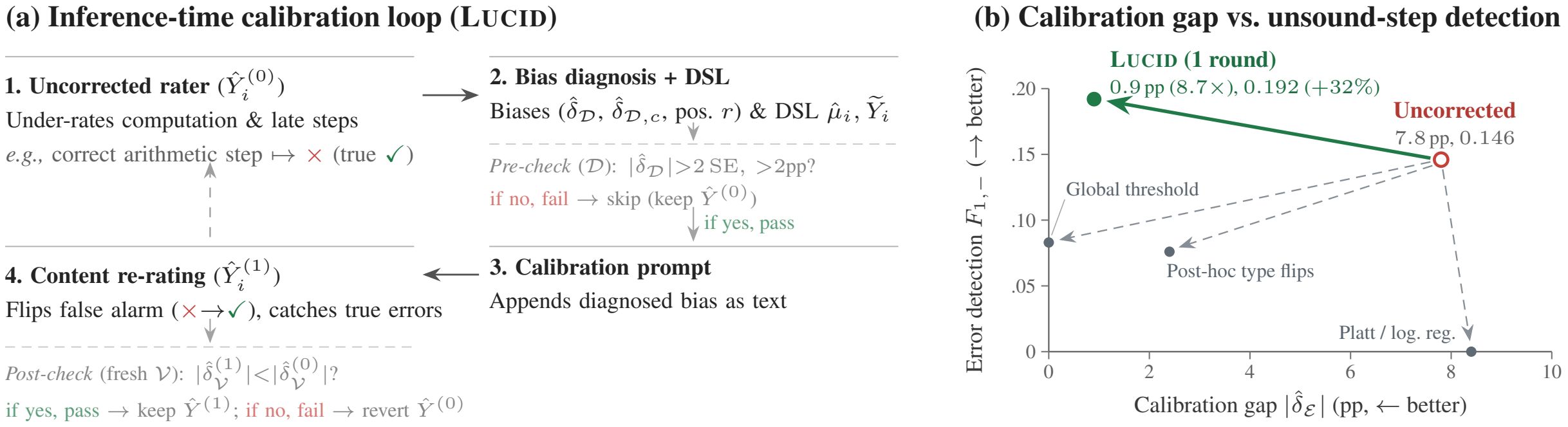

%% file: sections/01_introduction.tex
\section{Introduction}
\label{sec:introduction}

Chain-of-thought (CoT) traces often improve large language model (LLM) answer accuracy even when the intermediate reasoning is unsound: models can skip verification steps, make ungrounded logical leaps, or reach right answers for wrong reasons. Auditing reasoning quality---not just final answers---requires step-level judgment, yet expert annotation is expensive. Cheaper proxy LLM raters introduce systematic measurement error, a well-studied problem in computational social science \cite{grimmer2013text, gentzkow2019text}. This error propagates into downstream estimates unless the raters are \emph{calibrated}---meaning their predicted soundness rates (or probabilities) match expert rates overall and across subgroups \cite{keith2020text, geng2024surveyconfidence, shen2024universalcalibration, kapoor2024calibrationtuning, detommaso2024multicalibration}.

To correct this error without full expert annotation, we have an LLM such as GPT-5-mini predict binary step soundness across a corpus and use a small expert-labeled subset to debias downstream estimates---logistic regressions relating step features to soundness and sound-step transitions. Two estimators support this debiasing: Prediction-Powered Inference (PPI; \citealp{angelopoulos2023prediction}), which rectifies the regression score with expert-minus-proxy residuals, and Design-based Supervised Learning (DSL; \citealp{egami2023dsl}), which pairs a \emph{cross-fitted} outcome model---predicting expert soundness from step features and proxy ratings out-of-fold (holding out whole problems)---with inverse-probability-weighted expert residuals. \citet{de-pieuchon-etal-2025-benchmarking} demonstrate such debiasing for scalar document labels; we extend it to reasoning traces. A key question remains: \emph{once we also diagnose rater bias from that subset, can we feed that diagnosis back to improve the rater's judgments without extra labeling?}

We introduce \textsc{Lucid} (\textbf{L}oop for \textbf{U}nbiased \textbf{C}alibration via \textbf{I}nference-time \textbf{D}iagnosis). From expert-labeled steps, \textsc{Lucid} summarizes where the rater errs, appends that summary to the grading instructions, and prompts the rater to re-grade unseen steps (Figure~\ref{fig:showcase}a). On PRM800K \cite{lightman2024lets} (human step-level math labels), one \textsc{Lucid} pass shrinks GPT-5-mini's marginal calibration gap ($|\text{predicted} - \text{expert soundness rate}|$) 8.7-fold and raises unsound-step detection ($F_{1,-}$, the $F_1$ score on unsound steps) by 32\%. Purely prompt-based, it needs no retraining, stays stable across runs, and retains gains with 1{,}000 expert steps. We reuse that expert sample in DSL to debias regressions and calibrate corpus-wide probabilities; ranking traces with DSL's cross-fitted outcome model lifts top-1 (best-of-1) trajectory selection by 13.7pp---measured by whether the top-ranked trace yields the correct answer.

Our contributions are:
\begin{enumerate}
    \item \textbf{Prompt-based self-correction (\textsc{Lucid}).} We design a closed-loop procedure that diagnoses rater bias from expert labels, converts it into plain-text feedback, and prompts the rater to re-grade. On PRM800K, one pass cuts the calibration gap from 7.8pp to 0.9pp and raises $F_{1,-}$ by 32\% without retraining. Unlike prior heuristic-critique methods, \textsc{Lucid} grounds feedback in expert-labeled bias diagnosis, pairs re-grading with DSL debiasing, and guards deployment with pre-/post-checks that verify significant initial bias and roll back ineffective passes (Figure~\ref{fig:showcase}a; Section~\ref{sec:pref-correction}). Here, we find that one round captures most of the available gain (85.3\% $\pm$ 0.4pp), driven by the diagnosed calibration message.

    \item \textbf{Step-level bias diagnosis and DSL debiasing.} Extending \citet{de-pieuchon-etal-2025-benchmarking} to 594{,}690 PRM800K steps graded by GPT-5-mini, we first diagnose how the rater errs. On the held-out set, GPT-5-mini exhibits \emph{structured} signed bias (predicted minus expert soundness rate; negative $=$ under-rating): among six step types, it under-rates computation ($-14.6$pp) and verification ($-11.9$pp) most sharply, and its bias steepens in late positions (position--error correlation $r = -0.25$). Across four-run ablations, overall-gap feedback alone matches prompting with this structured profile, yet the profile still reveals whether errors reflect blanket strictness or concentrate in specific step types, and DSL uses its step features for corpus-wide calibration. On regressions at $n = 200$, DSL beats PPI and cuts coefficient root mean squared error ($\RMSE$, vs.\ all-expert fit) $7.2\times$ below expert-only fits, reaching $\RMSE < 0.5$ with $2.4\times$ fewer labels before expert-only fits catch up above $n \approx 1{,}000$.

    \item \textbf{Sound-step transitions and inference.} Addressing within-trace and within-problem dependence, we express consecutive sound-step type transitions as doubly-robust ratios of DSL-adjusted pair counts and cluster step-level regression standard errors by problem, raising them 3--3.5$\times$ at $n=1{,}000$.
\end{enumerate}

Finally, we delimit both the mechanisms and boundaries of inference-time feedback: it corrects threshold strictness (i.e., calibration) without improving class-balanced discrimination (i.e., sound-vs.-unsound separation under equal class weights). Consequently, this feedback helps conservative yet capable raters, fails when raters have skill deficits, and can over-correct already-calibrated raters. 

%% file: sections/02_related_work.tex
\section{Related Work}
\label{sec:related}

\paragraph{Statistical debiasing of LLM annotations.}
PPI \cite{angelopoulos2023prediction} rectifies estimating equations with expert residuals, preserving asymptotically valid confidence intervals under known sampling probabilities even with poor proxy predictions. Drawing on double/debiased ML \cite{chernozhukov2018double, bang2005doubly, kennedy2023towards}, DSL \cite{egami2023dsl} pairs out-of-fold (cross-fitted) outcome models predicting true labels from covariates and proxy predictions with doubly-robust pseudo-outcomes, remaining consistent if either sampling probabilities or the outcome model is right. While \citet{egami2023dsl} use honest forests \cite{wager2018estimation, athey2019generalized}, we use gradient-boosted trees predicting expert step soundness (Section~\ref{sec:step-soundness}). Recent extensions add cross-prediction-powered inference \cite{zrnic2024cross} and nonuniform sampling \cite{kluger2025prediction}. Benchmarking PPI and DSL across 16 dataset$\times$LLM pairs for logistic regression, \citet{de-pieuchon-etal-2025-benchmarking} show DSL lowers standardized RMSE ($\sRMSE$, coordinate-wise error normalized by true coefficient magnitude; see Appendix~\ref{app:background}) at small $n$. We extend these estimators to step ratings and study how rating errors distort within-trace transitions.

\paragraph{LLM self-correction and evaluation bias.}
In prompt-based self-correction frameworks (Self-Refine \cite{madaan2023selfrefine}, Reflexion \cite{shinn2023reflexion}), the model critiques its own outputs without statistical grounding, targeting generation quality rather than rating bias. In parallel, LLM judges exhibit systematic biases \cite{chang2024survey}: they show position and verbosity biases \cite{zheng2023judging}, favor their own generations \cite{panickssery2024llm}, and mimic human falsehoods \cite{lin2022truthfulqa}. Prior work mitigates judge bias through prompt design, aggregation (self-consistency, Platt scaling), or direct confidence calibration \cite{tian2023just}. Bridging these threads, \textsc{Lucid} diagnoses step-rating bias from expert labels, feeds it back under cluster-robust pre-/post-checks verifying initial bias and validation gain (Figure~\ref{fig:showcase}a; Section~\ref{sec:pref-correction}), and pairs re-grading with DSL debiasing.

\paragraph{Process supervision and agent evaluation.}
To support training process reward models (PRMs, step-level verifiers scoring intermediate reasoning), \citet{lightman2024lets} introduce PRM800K ($\sim$800K human MATH step labels) and \citet{wang-etal-2024-math} auto-generate step labels, while \citet{zheng-etal-2025-processbench} introduce ProcessBench to benchmark earliest-error identification. Rather than training verifiers, we use PRM800K and ProcessBench to diagnose rater bias and debias reasoning-quality estimates. Unlike trajectory-level agent benchmarks (AgentBoard \cite{ma2024agentboard}; SWE-bench \cite{jimenez2024swebench}), we debias step-level evaluation.

%% file: sections/03_methodology.tex
\section{Methodology}
\label{sec:methodology}

\subsection{\textsc{Lucid}: Prompt-Based Self-Correction}
\label{sec:pref-correction}

\paragraph{Procedure.}
Across $N$ reasoning steps (normalized trace position $\mathrm{pos}_i \in [0,1]$), the rater assigns initial ratings $\Yhat_i^{(0)} = \Yhat_i \in \{0,1\}$ and step types $s_i \in \{1,\dots,C\}$ ($C=6$ on PRM800K; Section~\ref{sec:step-soundness}), while experts label a diagnostic subset $\mathcal{D}$ ($|\mathcal{D}| = n_{\mathrm{diag}}$) with $Y_i \in \{0,1\}$. To avoid over-correcting calibrated raters or keeping failed passes (Section~\ref{sec:generalization}), each round $\tau \in \{0, 1, \dots\}$ runs three main phases guarded by two checks (Figure~\ref{fig:showcase}a). 

First, on $\mathcal{D}$, \textsc{Lucid} computes overall signed bias $\deltahat_{\mathcal{D}} = |\mathcal{D}|^{-1}\sum_{i\in\mathcal{D}}(\Yhat_i^{(\tau)} - Y_i)$ (negative $=$ under-rating; $|\deltahat_{\mathcal{D}}|$ is the marginal calibration gap) with cluster-robust $\mathrm{SE}$ (Section~\ref{sec:clustering}), per-type biases $\deltahat_{\mathcal{D},c} = |\mathcal{D}_c|^{-1}\sum_{i\in\mathcal{D}_c}(\Yhat_i^{(\tau)} - Y_i)$ over $\mathcal{D}_c = \{i \in \mathcal{D} : s_i = c\}$, and the Pearson correlation $r$ between $\mathrm{pos}_i$ and $\Yhat_i^{(\tau)} - Y_i$.

Second (\emph{Stage~1, pre-check}), \textsc{Lucid} proceeds only if $|\deltahat_{\mathcal{D}}| > 2\,\mathrm{SE}$ and $|\deltahat_{\mathcal{D}}| > 2$pp (else keeping $\Yhat^{(\tau)}$), formatting these statistics (without CIs) via a fixed plain-text template. 

Third, prompted with that feedback, the rater re-grades disjoint evaluation steps $\mathcal{E}$ ($n_{\mathrm{eval}} = 1{,}000$ on PRM800K; $500$ on multi-domain benchmarks) and a fresh expert-labeled validation batch $\mathcal{V}$ (100--500 steps) to produce $\Yhat^{(\tau+1)}$. 

Finally (\emph{Stage~2, post-check}), \textsc{Lucid} keeps $\Yhat^{(\tau+1)}$ only if $|\deltahat_{\mathcal{V}}^{(\tau+1)}| < |\deltahat_{\mathcal{V}}^{(\tau)}|$, rolling back to $\Yhat^{(\tau)}$ otherwise. Here, we find that one round captures most of the available gain (Section~\ref{sec:iterative-prompt-results}).

\paragraph{Feedback format.}
When Stage~1 passes, \textsc{Lucid} populates a three-part plain-text bias summary; to benchmark all-$N$ diagnosis before restricting to $\mathcal{D}$ on PRM800K, our canonical prompt states \emph{overall calibration}, \textit{``You predict 83.1\% of steps correct, but the true rate is 91.3\% (you under-rate by 8.2pp overall)''}; \emph{per-type biases}, \textit{``You under-rate computation steps by 14.8pp and verification steps by 14.5pp''}; and \emph{position trend}, \textit{``You also under-rate later steps, $r = -0.22$''}, closing with a directive to judge late computation and verification steps less conservatively. Because uncorrected bias is stable across subsets (Section~\ref{sec:results}), diagnosing on disjoint $\mathcal{D}$ ($n_{\mathrm{diag}} = 1{,}000$) matches full-corpus gains (Tables~\ref{tab:n-diag-sensitivity}, \ref{tab:label-noise}, Appendix~\ref{app:self-correction-sensitivity}).

\subsection{Step-Level Soundness Framework}
\label{sec:step-soundness}

Alongside re-grading on an unlabeled corpus, we reuse $\mathcal{D}$ with $\Yhat$ to \emph{predict} calibrated expert soundness $Y_i \in \{0,1\}$ across all $N$ steps and \emph{debias} regressions relating step features to soundness (and sound-step transitions on pair-labeled samples; Section~\ref{sec:transition}). Each step $i \in \{1,\dots,N\}$ has expert label $Y_i \in \{0,1\}$ on $\mathcal{D}$ plus proxy LLM rating $\Yhat_i \in \{0,1\}$ and step type $s_i$ (Setup, Computation, Inference, Verification, Strategy, Restatement; Table~\ref{tab:taxonomy}). We form $X_i \in \R^{p+1}$ from $s_i$, $\mathrm{pos}_i$, log token length $\mathrm{len}_i$, difficulty, and an intercept (Appendix~\ref{app:features}). Our debiasing target is the logistic parameter $\theta \in \R^{p+1}$ solving $\E[u(Y_i, X_i; \theta)] = 0$ for score $u(Y_i, X_i; \theta) = X_i(Y_i - \sigma(X_i^\top \theta))$ ($\sigma$: sigmoid).

Both PPI and DSL debias $\theta$ using $(X_i, \Yhat_i)$ across all $N$ steps and $Y_i$ on $\mathcal{D}$ ($|\mathcal{D}| = n$, matching $n_{\mathrm{diag}}$ when shared), where $b_i = \1\{i \in \mathcal{D}\}$ and $\pi_i = \Pr(b_i = 1 \mid X_i, \Yhat_i) = n/N$. PPI adds the mean labeled residual on $\mathcal{D}$ to the plug-in score. DSL fits a two-fold \emph{cross-fitted} gradient-boosted outcome model $\hat{\mu}_i = \hat{\mu}(X_i, \Yhat_i)$ predicting expert soundness $\mu(X_i, \Yhat_i) = \E[Y_i \mid X_i, \Yhat_i]$---holding out whole problems per fold so $\hat{\mu}_i$ is an out-of-fold probability (also used for corpus calibration and ranking)---forms doubly-robust pseudo-outcomes $\Ytilde_i = \hat{\mu}_i + \frac{b_i}{\pi_i}(Y_i - \hat{\mu}_i)$, and solves $\sum_{i=1}^N u(\Ytilde_i, X_i; \theta) = 0$ (Appendix~\ref{app:background}). Against all-$N$ expert fit $\thetaref$, we evaluate expert-only $\thetahat_{\mathrm{exp}}$ (on $\mathcal{D}$), uncorrected imputation $\thetahat_{\mathrm{imp}}$ (on all $\Yhat_i$), PPI, and DSL via $\RMSE(\thetahat) = (|\mathcal{J}|^{-1}\sum_{j\in\mathcal{J}}(\thetahat_j - \thetaref_j)^2)^{1/2}$ over identifiable coordinates $\mathcal{J}$ (Appendix~\ref{app:background}).

\subsection{Transitions Between Sound Steps}
\label{sec:transition}

To characterize valid reasoning flow, we estimate how often consecutive sound steps transition from type $c$ to $c'$. Index the $M$ consecutive within-trace pairs by $\ell \in \{1,\dots,M\}$, linking $(i(\ell), i(\ell)+1)$ with LLM-assigned step types $(s_{\ell,1}, s_{\ell,2}) = (s_{i(\ell)}, s_{i(\ell)+1}) \in \{1,\dots,C\}^2$ and joint expert soundness $A_\ell = Y_{i(\ell)} Y_{i(\ell)+1} \in \{0,1\}$. Pairs at different trace positions need not share a distribution, so we define the target through the limiting share of sound $c$-to-$c'$ pairs among the $M$ consecutive pairs, $\lambda_{cc'} = \lim_{M\to\infty}\frac{1}{M}\sum_{\ell=1}^{M}\Pr(s_{\ell,1}=c, s_{\ell,2}=c', A_\ell=1)$ (assumed to exist):
\begin{equation}
    T_{cc'} = \frac{\lambda_{cc'}}{\sum_{c''=1}^{C}\lambda_{cc''}}.
    \label{eq:sound-transition}
\end{equation}
For identically distributed pairs, $T_{cc'} = \Pr(s_{\ell,2}=c' \mid s_{\ell,1}=c, A_\ell=1)$. We form $\Tref$ by normalizing sound-pair counts ($A_\ell = 1$) within each starting type, and $\That_{\mathrm{uncorr}}$ (pair-level analogue of $\thetahat_{\mathrm{imp}}$) using predicted sound pairs ($\Ahat_\ell = \Yhat_{i(\ell)}\Yhat_{i(\ell)+1} = 1$). Parallel to $(b_i, \pi_i, \mu)$, let $b_{A,\ell} \in \{0,1\}$ indicate expert labels on both endpoints of pair $\ell$, and let $W_\ell$ collect observed pair features ($s_{\ell,1}, s_{\ell,2}$, both predictions, and sampling covariates). We also define $\pi_A(W_\ell)=\Pr(b_{A,\ell}=1\mid W_\ell)$ (abbreviated $\pi_{A,\ell}$) and $\mu_A(W_\ell)=\E[A_\ell\mid W_\ell]$. We cross-fit $\hat{\mu}_{A,\ell} = \hat{\mu}_A(W_\ell)$ by problem and set $\hat{\pi}_{A,\ell}$ to the known design (or cross-fitted) probability. Following \citet[Proposition~2]{egami2023dsl} and \citet{bang2005doubly}, we then construct the pair pseudo-outcome $\Atilde_\ell = \hat{\mu}_{A,\ell} + \frac{b_{A,\ell}}{\hat{\pi}_{A,\ell}}(A_\ell-\hat{\mu}_{A,\ell})$ and estimate $T_{cc'}$ as the ratio of summed pseudo-outcomes for $c$-to-$c'$ pairs to all pairs leaving $c$:
\begin{equation}
    \That_{cc'} =
    \frac{\sum_{\ell:s_{\ell,1}=c}\1\{s_{\ell,2}=c'\}\,\Atilde_\ell}
         {\sum_{\ell:s_{\ell,1}=c}\Atilde_\ell}.
    \label{eq:transition-estimate-main}
\end{equation}
Under the sampling and regularity conditions in Appendix~\ref{app:sound-pairs}, $\That_{cc'}$ is consistent if either the pair inclusion probabilities or the conditional probability that both steps are sound is correctly specified. When pairs are sampled uniformly for expert review, their inclusion probabilities are known by design.

\subsection{Clustering-Aware Inference}
\label{sec:clustering}

Because steps and traces from the same problem share errors, naive standard errors understate uncertainty in both our pre-check (Section~\ref{sec:pref-correction}) and step-level regressions (Section~\ref{sec:step-soundness}). Summing per-step score contributions $\psi_i(\thetahat)$ (e.g., $\psi_i(\theta) = u(\Ytilde_i, X_i; \theta)$ for DSL; Appendix~\ref{app:background}) within each problem $g \in \{1,\dots,G\}$ gives $S_g=\sum_{i\in g}\psi_i(\thetahat)$. With summed-score Jacobian $\hat H$ on identifiable coordinates, the sandwich estimator \citep{liang1986longitudinal} gives
\begin{equation}
    \hat V=\hat H^{-1}\left(\sum_{g=1}^G S_gS_g^\top\right)\hat H^{-\top}.
    \label{eq:cluster-robust-se}
\end{equation}
The large-sample calculation assumes independent problems, finite score variances, a nonsingular limit for the normalized Jacobian, and enough clusters that no single problem dominates. For scores with a fitted outcome model, keeping whole problems in each cross-fitting fold separates training and evaluation clusters \citep{chiang2022cluster}.

\paragraph{Summary.}
In theory, known inclusion probabilities secure doubly-robust consistency for DSL regressions and sound-step transitions even under imperfect outcome models (Proposition~\ref{prop:sound-pairs}), while problem-grouped cross-fitting, cluster-robust SEs, and two-stage gating guard against within-problem dependence and over-correction. Section~\ref{sec:experiments} details the benchmarks and protocols testing finite-sample gains. 

%% file: sections/04_experiments.tex
\section{Experiments}
\label{sec:experiments}

\paragraph{Data.}
We use PRM800K \cite{lightman2024lets}: $\sim$800K human step-level correctness labels (on a $\{-1, 0, +1\}$ incorrect/neutral/correct scale) for MATH \cite{hendrycks2021math} solutions. After filtering and binarizing ($Y_i = 1$ if rating $= +1$, else $Y_i = 0$), we obtain $N = 594{,}690$ steps across 90{,}340 traces and $G = 11{,}267$ problems. GPT-5-mini \cite{openai2025gpt5} grades and classifies all $N$ steps at temperature 0, and we test cross-model transfer on GLM-5.2 \cite{zai2026glm52} (Appendix~\ref{app:self-correction-details}). For cross-dataset and multi-domain transfer, we evaluate on ProcessBench \cite{zheng-etal-2025-processbench}, EntailmentBank \cite{dalvi2021explaining}, Mind2Web \cite{deng2023mind2web}, and ProofWriter \cite{tafjord2021proofwriter}.

\paragraph{Features.}
Each step has $p = 9$ covariates ($X_i \in \R^{10}$ with intercept): step type (5 one-hot dummies for the $C = 6$ categories), normalized position, log step length, and difficulty (2 dummies; see Appendix~\ref{app:features}). An augmented set ($p = 24$, $X_i \in \R^{25}$) adds 15 cluster indicators from $k$-means ($k{=}15$) on \texttt{all-MiniLM-L6-v2} step-and-problem embeddings, testing whether grouping semantically similar steps helps DSL pool bias structure at small $n$ (Appendix~\ref{app:adaptive}).

\paragraph{Design.}
For the regression comparison, we fix $N=594{,}690$, vary the expert budget $n$ from 200 to approximately $N$ over 10 log-spaced values (30 repetitions per budget), and compare five estimators (Table~\ref{tab:methods}) by coefficient $\RMSE$ and standard errors ($\sRMSE$ is defined in Appendix~\ref{app:background}). For prompt self-correction on PRM800K, we use the corpus diagnostic profile for the canonical prompt, verify subset-only diagnosis at $n_{\mathrm{diag}} \in \{500, 1{,}000, 2{,}000\}$ (Appendix~\ref{app:self-correction-sensitivity}), and evaluate on $n_{\mathrm{eval}} = 1{,}000$ disjoint steps $\mathcal{E}$. On the multi-domain benchmarks, we set $n_{\mathrm{diag}} = n_{\mathrm{eval}} = 500$. For transitions, we draw $m=1{,}000$ consecutive pairs uniformly without replacement, observe expert labels on both endpoints, and estimate $\That_{\mathrm{DSL}}$ using the known $\pi_{A,\ell} = m/M$ ($\hat{\pi}_{A,\ell} = \pi_{A,\ell}$). We then compare consecutive-pair matrices against $\Tref$ via relative Frobenius error $\|\That - \Tref\|_F / \|\Tref\|_F$ (Appendix~\ref{app:transition-matrix}).

%% file: sections/05_results.tex
\section{Results}
\label{sec:results}

Because 91.6\% of evaluation steps are sound, raw accuracy alone misleads: an always-positive predictor reaches 91.6\% while catching no errors ($F_{1,-} = 0$). Alongside step accuracy (Table~\ref{tab:headline}), we therefore report the marginal calibration gap ($|\deltahat_{\mathcal{E}}| = |\mathcal{E}|^{-1}\big|\sum_{i\in\mathcal{E}}(\Yhat_i - Y_i)\big|$) and unsound-step $F_{1,-}$ (treating unsound steps $Y_i = 0$ as positive; Table~\ref{tab:calibration}). For downstream evaluation, we rank candidate trajectories with DSL's cross-fitted step probabilities.

\begin{table}[h]
\centering
\resizebox{\columnwidth}{!}{%
\begin{tabular}{lrrrc}
\toprule
\textbf{Setting} & \textbf{Original} & \textbf{Corrected} & $\boldsymbol{\Delta}$ & \textbf{Cost} \\
\midrule
PRM800K (GPT-5-mini) & 79.0\% & 85.7\% & $+$6.7pp & \$0.30 \\
PRM800K (GLM-5.2) & 84.4\% & 87.5\% & $+$3.1pp & \$0.30 \\
ProcessBench & 81.4\% & 86.6\% & $+$5.2pp & \$0.30 \\
\midrule
DSL $\RMSE$ ($n{=}200$) & 6.09 & 0.84 & $-$5.25 & --- \\
\bottomrule
\end{tabular}%
}
\caption{Headline results ($\Delta = \text{Corrected} - \text{Original}$). One round of self-correction closes the PRM800K calibration gap $8.7\times$ (7.8pp$\to$0.9pp; signed bias $-$7.8pp$\to-$0.9pp) at \$0.30 API cost. Because 91.6\% of evaluation steps are sound, shifting the rater's threshold raises raw accuracy substantially, whereas accuracy on a class-balanced subset moves only from 57.1\% to 58.3\% (Appendix~\ref{app:balanced-ablation}). This contrast shows that threshold calibration drives the raw-accuracy gain. At $n{=}200$, DSL cuts regression $\RMSE$ from 6.09 (Expert-only) to 0.84 ($7.2\times$, mainly at $n \lesssim 1{,}000$; Section~\ref{sec:dsl-results}).}
\label{tab:headline}
\end{table}

\subsection{\textsc{Lucid}: Prompt-Based Self-Correction}
\label{sec:iterative-prompt-results}

On PRM800K, one \textsc{Lucid} pass cuts GPT-5-mini's calibration gap nearly ninefold and lifts unsound-step detection ($F_{1,-}$) by nearly a third (Tables~\ref{tab:headline} and~\ref{tab:calibration}; cross-benchmark summary in Table~\ref{tab:calibration-headline}; worked trace in Table~\ref{tab:qualitative-example}). Across four runs, \textsc{Lucid} yields stable step-accuracy gains and converges in one pass, with a second round adding little (Table~\ref{tab:iterative-prompt}). On an independent 5{,}000-step evaluation set, \textsc{Lucid} reproduces both the accuracy gain and the concentration of improvements in computation and verification (Table~\ref{tab:iterative-prompt}). At \$0.30 in API calls atop the shared expert budget, \textsc{Lucid} beats expert-only estimation at matched labeling cost and---unlike fine-tuned verifiers---avoids retraining when rater behavior shifts (Table~\ref{tab:cost}).

\begin{figure}[t]
    \centering
    \includegraphics[width=0.92\columnwidth]{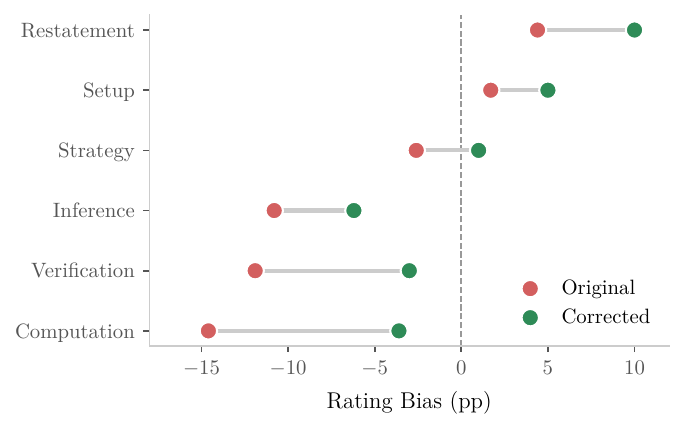}
    \caption{Self-correction pulls severe per-type rating biases (red) toward zero (green) on PRM800K, led by computation ($-$14.6$\to-$3.6pp) and verification ($-$11.9$\to-$3.0pp).}
    \label{fig:bias-dumbbell}
\end{figure}

\begin{table}[t]
\centering\small
\begin{tabular}{lcc}
\toprule
\textbf{Metric} & \textbf{Original} & \textbf{Corrected} \\
\midrule
Calibration gap (overall) $\downarrow$ & 7.8pp & 0.9pp \\
Computation gap $\downarrow$ & 14.6pp & 3.6pp \\
Verification gap $\downarrow$ & 11.9pp & 3.0pp \\
Inference gap $\downarrow$ & 10.8pp & 6.2pp \\
Unsound detection $F_{1,-}$ $\uparrow$ & 0.146 & 0.192 \\
\bottomrule
\end{tabular}
\caption{Calibration gap ($|\deltahat_{\mathcal{E}}|$) and error-detection metrics on PRM800K ($n_{\mathrm{eval}}{=}1{,}000$). Self-correction cuts the gap $8.7\times$ overall and up to $4.1\times$ per type while lifting $F_{1,-}$ by $+$0.046 ($+$32\%).}
\label{tab:calibration}
\end{table}

\paragraph{Per-type and position improvements.}
Uncorrected GPT-5-mini is well calibrated on early framing steps (setup, strategy) and early positions, but grows monotonically stricter across a trace---from $+$2.7pp in the first quintile to $-$32.2pp in the final quintile (59.9\% predicted vs.\ 92.1\% true; peaking at $-$43.9pp on late restatement and $-$34.4pp on late computation; Figure~\ref{fig:bias-dumbbell}; Table~\ref{tab:calibration}; Appendix~\ref{app:bias-analysis}; Figure~\ref{fig:calibration-by-type}). Three factors drive this late-step conservatism: skepticism accumulates over longer context chains despite $\approx$92\% true soundness, the rater conflates step soundness with final-answer proximity, and later steps are longer and more complex ($>$200-char steps show $\approx-$30pp bias). \textsc{Lucid}'s gains track this profile: the most under-rated types and final quintile improve most while well-calibrated categories move little (Table~\ref{tab:iterative-prompt-per-type}). Initial under-rating correlates positively with post-feedback gain across the six types in both the primary set and 5{,}000-step replication (though six types limit slope precision), and \textsc{Lucid} fixes roughly two-thirds of initial mistakes in the final quintile versus under a third in the first (Table~\ref{tab:iterative-prompt-per-type}; Appendix~\ref{app:bias-analysis}).

\begin{figure}[t]
    \centering
    \includegraphics[width=0.92\columnwidth]{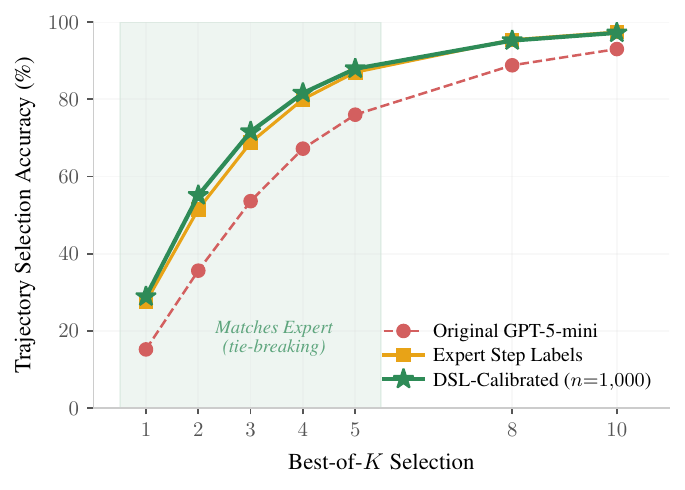}
    \caption{Best-of-$K$ selection on 7{,}348 MATH problems. DSL-calibrated scores ($n{=}1{,}000$, green) beat GPT-5-mini by $+$13.7pp ($K{=}1$) and $+$19.5pp ($K{=}2$), matching Expert labels at $K \le 5$.}
    \label{fig:teaser}
\end{figure}

\paragraph{Downstream validation: trajectory filtering.}
To test downstream selection across 7{,}348 MATH problems, we rank each problem's candidate traces by mean step score and measure best-of-$K$ (top-$K$) accuracy: whether any top-$K$ trace gets the right answer. \textsc{Lucid}'s prompt loop re-rates only $1{,}000$ sampled steps in binary form, whereas corpus-wide ranking requires scoring every trace and benefits from continuous tie-breaking scores. We therefore score all steps with DSL's cross-fitted outcome model. These DSL-calibrated probabilities nearly double top-1 accuracy over raw GPT-5-mini ratings and match expert labels through best-of-5 (Figure~\ref{fig:teaser}; full $K = 1\text{--}10$ curve in Table~\ref{tab:trajectory-filtering-full}).

\paragraph{Why DSL can match the expert oracle.}
Two factors explain why DSL-calibrated proxy scores match the expert oracle in Figure~\ref{fig:teaser}. First, we evaluate trajectory selection by final-answer correctness, yet solutions with flawed intermediate steps can still reach the right answer. Unweighted averages of binary step labels therefore provide only a coarse ranking signal. Second, while the expert oracle uses ground-truth corpus step labels (unavailable in practice), DSL's cross-fitted outcome model approximates them from the diagnostic sample while supplying continuous probabilities that break ties among traces.

\paragraph{Downstream PRM transfer.}
We also train lightweight feature-based PRMs on raw GPT-5-mini ratings, DSL outcome probabilities ($\hat{\mu}_i$), and expert labels (Table~\ref{tab:prm-transfer}); 0.5-thresholding collapses all three to the majority class. With continuous probabilities, the DSL-calibrated PRM halves expected calibration error (ECE) and lifts top-1 selection by 1.8pp over the raw-label PRM (ECE measures binned probability-vs.-rate error, unlike the binary gap).

\paragraph{Baselines.}
Returning to binary step re-grading on our skewed evaluation set, accuracy-maximizing calibrators (Platt scaling, logistic regression [LR], and MiniLM) hit the majority-class ceiling by predicting every step sound and catching no errors (Figure~\ref{fig:showcase}b; Table~\ref{tab:baselines}). Rate-adjustment baselines apply the diagnosed bias without re-reading text, using either a global threshold shift (matching the overall soundness rate) or post-hoc random flips matched by step type and position. These adjustments close the calibration gap and match or beat \textsc{Lucid}'s raw accuracy. However, because both content-blind adjustments discard true error detections alongside false alarms, they cut unsound-step $F_{1,-}$ nearly in half. With the same bias summary, \textsc{Lucid} re-grades each step on content, halving false alarms while preserving unsound-step recall. Among the inference-time methods in Table~\ref{tab:baselines}, only \textsc{Lucid} improves both calibration and error detection.

Still, on a class-balanced subsample with equal sound and unsound counts, \textsc{Lucid}'s accuracy and $F_{1,-}$ remain within noise (Appendix~\ref{app:balanced-ablation}; balanced accuracy $(\mathrm{Recall}_+ + \mathrm{Recall}_-)/2$ in Table~\ref{tab:baselines}). This invariance confirms that \textsc{Lucid} calibrates the decision threshold without sharpening discrimination. 

By contrast, detection-oriented Qwen3.5-LoRA \cite{qwen2026qwen35, hu2021lora} verifiers (0.8B--9B) reach slightly higher $F_{1,-}$ and balanced accuracy at smaller scales at lower overall accuracy (Table~\ref{tab:baselines}). Prompt self-correction thus fits closed-weight APIs and shifting biases, whereas fine-tuning remains stronger with open weights for balanced discrimination.

\paragraph{Ablation.}
Across four runs per condition, ablating the feedback summary shows that overall-gap, per-type, and full three-part prompts each lift accuracy by roughly 5--6pp over no feedback with overlapping paired-bootstrap CIs, whereas a position-only prompt gains less because positional bias is weaker alone (Table~\ref{tab:ablation-app}). Detailed, concise, and minimal wordings likewise perform indistinguishably (Table~\ref{tab:wording-sensitivity}). Although overall-gap feedback suffices for prompt re-grading, the structured profile still shows whether errors reflect blanket strictness or concentrate in specific hard step types (Section~\ref{sec:generalization}). In turn, DSL uses that profile's underlying step features $X_i$ to calibrate corpus-wide probabilities.

\subsection{Generalization}
\label{sec:generalization}

\begin{table*}[t]
\centering\small
\setlength{\aboverulesep}{0.2ex}%
\setlength{\belowrulesep}{0.35ex}%
\renewcommand{\arraystretch}{0.92}%
\begin{tabular}{@{}lrrrrrc@{}}
\toprule
 & \multicolumn{3}{c}{\textbf{Step Accuracy}} & \multicolumn{2}{c}{\textbf{Calib.\ Gap ($|\deltahat_{\mathcal{E}}|$)}} & \\
\cmidrule(lr){2-4}\cmidrule(lr){5-6}
\textbf{Domain} & \textbf{Orig} & \textbf{Corr} & $\boldsymbol{\Delta}$ & \textbf{Orig} & \textbf{Corr} & \textbf{Rule} \\
\midrule
PRM800K (math) \cite{lightman2024lets} & 79.0\% & 85.7\% & $+$6.7pp & 7.8pp & 0.9pp & apply, keep \\
EntailmentBank (science) \cite{dalvi2021explaining} & 94.2\% & 96.6\% & $+$2.4pp & 5.8pp & 3.4pp & apply, keep \\
Mind2Web (agents, calibration) \cite{deng2023mind2web} & 10.8\% & 97.4\% & $+$86.6pp & 89.2pp & 2.6pp & apply, keep \\
Mind2Web (agents, balanced) \cite{deng2023mind2web} & 50.8\% & 61.0\% & $+$10.2pp & 0.4pp & 38.6pp & \textbf{skip} \\
ProofWriter (logic) \cite{tafjord2021proofwriter} & 94.4\% & 94.0\% & $-$0.4pp & 4.8pp & 4.8pp & apply, roll back \\
\bottomrule
\end{tabular}
\caption{Multi-domain generalization. \textsc{Lucid} transfers to calibration bias but over-corrects calibrated raters and fails on competence deficits; the two-stage deployment rule (Section~\ref{sec:pref-correction}) skips balanced Mind2Web (Stage 1) and rolls back ProofWriter (Stage 2) while keeping all three beneficial corrections.}
\label{tab:multidomain}
\end{table*}

\paragraph{Cross-model transfer.}
Applying feedback diagnosed from GPT-5-mini's ratings directly to GLM-5.2 \cite{zai2026glm52} still improves step accuracy on PRM800K (Table~\ref{tab:headline}; Table~\ref{tab:cross-model-sc}). GLM-5.2 starts much closer to overall calibration than GPT-5-mini yet shares the tendency to under-rate computation and verification steps, so its smaller gain tracks its milder initial bias (Table~\ref{tab:bias-profile-cross-model}).

\paragraph{Cross-dataset transfer (ProcessBench).}
On ProcessBench (16{,}548 steps across GSM8K, MATH, OlympiadBench, and OmniMath), one \textsc{Lucid} pass (with the same feedback template) lifts GPT-5-mini's accuracy by over five percentage points and beats few-shot baselines across $k = 1\text{--}10$ examples (Tables~\ref{tab:headline}, \ref{tab:processbench}, and~\ref{tab:fewshot-comparison}). Across splits, gains track initial miscalibration: accuracy rises over ten points on worst-calibrated OmniMath but slips slightly on near-ceiling GSM8K (Table~\ref{tab:processbench}).

\paragraph{Multi-domain.}
Across three non-mathematical benchmarks with domain-specific taxonomies ($n_{\mathrm{diag}} = n_{\mathrm{eval}} = 500$; Table~\ref{tab:multidomain}), \textsc{Lucid} helps wherever a capable rater grades too strictly: it cuts EntailmentBank's calibration gap by 41\% and lifts Mind2Web gold trajectories (all valid, 10.8\% initial acceptance) to 97.4\%, confirming threshold strictness as a key bottleneck. Conversely, unchecked feedback severely over-corrects calibrated balanced Mind2Web (0.4pp initial gap) and leaves ProofWriter unchanged where errors reflect logical derivation skill deficits (Table~\ref{tab:boundary-profiles}). Without tuning, the two-stage rule (Section~\ref{sec:pref-correction}; Table~\ref{tab:multidomain}, ``Rule'') separates these regimes: though $\deltahat_{\mathcal{D}} = -5.2$pp on both balanced Mind2Web and EntailmentBank, cluster-robust SE is 3.4pp ($|\deltahat_{\mathcal{D}}| < 2\,\mathrm{SE}$, skipped at Stage~1) versus 1.1pp ($|\deltahat_{\mathcal{D}}| > 2\,\mathrm{SE}$, admitted), while Stage~2 rolls back ProofWriter when $|\deltahat_{\mathcal{V}}|$ fails to shrink.

\subsection{DSL Debiasing and Diagnostic Robustness}
\label{sec:dsl-results}

\begin{figure}[t]
    \centering
    \includegraphics[width=0.92\columnwidth]{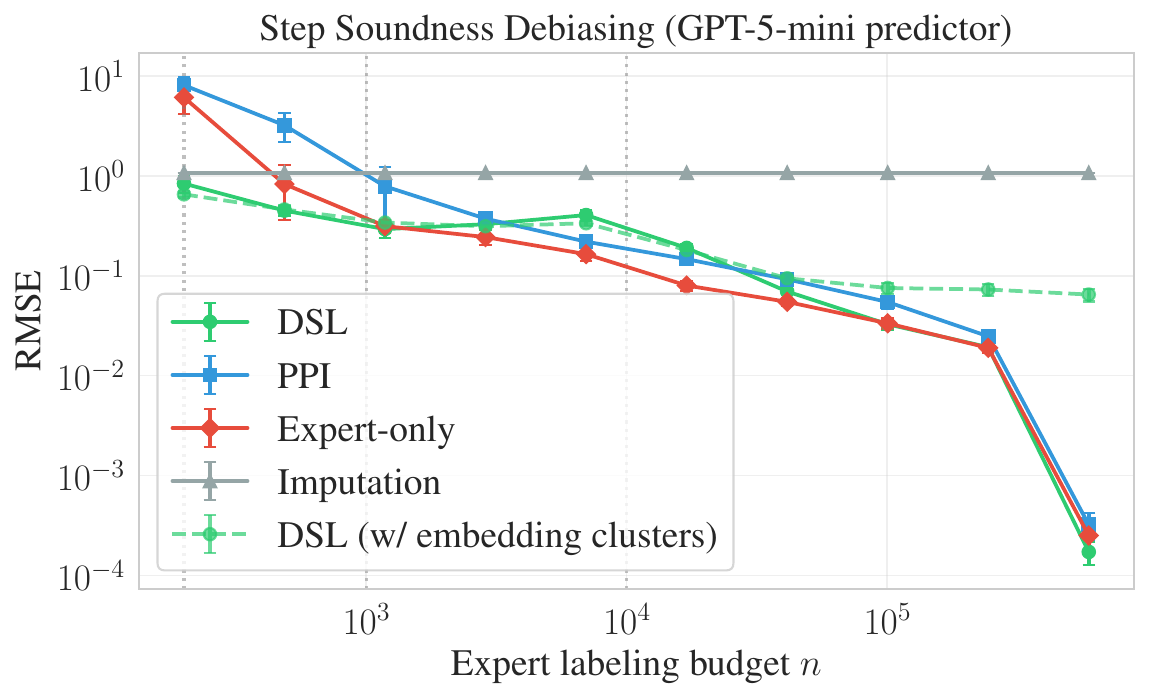}
    \caption{Coefficient $\RMSE$ vs.\ expert budget $n$ (GPT-5-mini predictor; 95\% CIs over 30 runs). DSL dominates at small $n$.}
    \label{fig:vary-expert}
\end{figure}

\paragraph{Sample efficiency across expert budgets.}
Turning from prompt transfer to PRM800K step-level regressions (Section~\ref{sec:step-soundness}), DSL cuts coefficient $\RMSE$ over $7\times$ at $n = 200$ vs.\ expert-only fits, beats PPI, and reaches $\RMSE < 0.5$ with under half the labels (Figure~\ref{fig:vary-expert}; Table~\ref{tab:stat-tests}). Near $n \approx 1{,}200$, both match; expert-only modestly edges out our fixed GBDT at $n \approx 2{,}900\text{--}17{,}000$ before reconverging at large $n$ (Table~\ref{tab:adaptive-app}).

\paragraph{Transition matrices and clustered inference.}
Within-trace and within-problem dependence affects both point estimates and uncertainty. For consecutive pairs, $\That_{\mathrm{uncorr}}$ undercounts transitions into computation and inference and overcounts those into strategy relative to $\Tref$ (Figure~\ref{fig:transition-bias-app}), while DSL cuts relative Frobenius error by about two-thirds (Figure~\ref{fig:transition-heatmap}; Appendix~\ref{app:transition-matrix}). Problem-clustered SEs triple independent-step SEs across regression estimators (Table~\ref{tab:clustering}).

\paragraph{Diagnostic robustness.}
Beyond DSL, our diagnosis and prompt loop degrade gracefully under taxonomy and label noise, respond selectively to structured bias, and transfer across same-domain problems (Appendix~\ref{app:self-correction-sensitivity}). For step-type noise, GLM-5.2 agrees with GPT-5-mini on computation and strategy yet diverges on verification and restatement (Table~\ref{tab:taxonomy-agreement}); even under 30\% random step-type reassignment, our diagnosis preserves every category's sign (Table~\ref{tab:noise-controls}). For expert-label noise, corrupting 10\% of diagnostic labels weakens \textsc{Lucid}'s gain without reversing it (Table~\ref{tab:label-noise}). For signal selectivity, when we replace structured rater errors with random flips, \textsc{Lucid} leaves accuracy essentially flat, confirming that the loop ignores unstructured noise (Table~\ref{tab:noise-controls}). Finally, profiles diagnosed on half the problems transfer to held-out same-domain problems, though a single dataset-wide prompt (not tailored by trace position) still under-corrects late steps (Table~\ref{tab:staleness}).

%% file: sections/06_conclusion.tex
\section{Conclusion}
\label{sec:conclusion}

\textsc{Lucid} closes the calibration loop at inference time: it diagnoses rater bias from expert-labeled steps, converts it into plain-text feedback, and prompts re-grading without retraining or weight access. On PRM800K, one pass cuts GPT-5-mini's calibration gap nearly ninefold and lifts unsound-step $F_{1,-}$ by nearly a third, transferring across models and domains. Meanwhile, balanced ablations confirm that prompt feedback calibrates strictness without sharpening discrimination. Accordingly, \textsc{Lucid}'s pre-/post-checks admit re-grading for strict capable raters, skip calibrated ones, and roll back passes under skill deficits. Beyond re-grading, reusing that expert set with raw ratings in DSL lifts best-of-$K$ selection via cross-fitted probabilities to match expert labels, cuts regression $\RMSE$ up to $7.2\times$ at $n \lesssim 1{,}000$, and (on pair-labeled steps) cuts the error of sound-step transition estimation by nearly two-thirds. 

%% file: sections/07_limitations.tex
\section{Limitations}
\label{sec:limitations}

\paragraph{Calibration ceiling and downstream scope.}
\textsc{Lucid} loosens a rater's threshold without sharpening the rater's separation between sound and flawed steps. As a result, the rater still triggers residual false alarms (sound steps rated unsound, i.e., soundness false negatives) on computation and inference, keeping overall accuracy below the 91.6\% majority-class ceiling. Trivial calibration baselines therefore win on raw accuracy by predicting every step sound while catching no errors (Section~\ref{sec:iterative-prompt-results}). When practitioners have open model weights and prioritize balanced-class discrimination, supervised fine-tuning outperforms prompt self-correction (Table~\ref{tab:baselines}). 

Regarding our deployment rule, although our pre- and post-correction checks skip calibrated raters and revert ineffective passes, we cannot yet distinguish rater conservatism from missing skill \emph{before} running a re-grading pass. 

Finally, on downstream scope, we rank candidate traces in our best-of-$K$ experiment with DSL's cross-fitted outcome model rather than prompt-corrected ratings; evaluating trajectory selection on \textsc{Lucid}'s re-ratings directly will require re-grading all candidate traces.

\paragraph{Evaluator coverage and run variance.}
We evaluate two model families, GPT-5-mini and GLM-5.2 (Section~\ref{sec:generalization}); other model families such as Claude or DeepSeek may show different biases or prompt sensitivity. In addition, outside the four-run PRM800K replications and ablations (Table~\ref{tab:iterative-prompt}), benchmark accuracies come from single temperature-0 passes and carry minor API run-to-run variance.

\paragraph{Reference labels and step taxonomy.}
Both \textsc{Lucid} and DSL calibrate toward the human reference (here, PRM800K annotations) instead of objective soundness. Moderate label noise only attenuates \textsc{Lucid}'s correction (Appendix~\ref{app:label-noise}), whereas systematic annotator skew transfers directly into the diagnosed profile, where the diagnostic pipeline cannot detect it from within. We outline untested multi-pool and noise-aware mitigations in Section~\ref{sec:ethical-considerations}. Similarly, we defined the six-category step taxonomy and applied it with GPT-5-mini without human inter-annotator validation. Because GLM-5.2 agrees closely with GPT-5-mini on computation and strategy steps yet diverges on verification and restatement (Appendix~\ref{app:taxonomy-agreement}), human validation and finer or continuous step features remain important next steps.

\paragraph{Profile shift, unstructured noise, and sample budgets.}
A single dataset-wide bias profile under-corrects severe late-position bias and is subject to profile staleness---outdated bias summaries losing calibration accuracy when reused across shifted step distributions, new domains, or prompt revisions without re-diagnosis---pointing toward position-adaptive feedback and nonuniform or multi-model extensions \cite{kluger2025prediction, zrnic2024cross} (Appendix~\ref{app:staleness}). Prompt self-correction also breaks down when rater errors are unstructured noise, when smaller models ignore complex prompt directives, or when scarce diagnostic labels ($n_{\mathrm{diag}} < 500$) prevent reliable per-type rate estimation (Appendix~\ref{app:self-correction-sensitivity}). DSL exhibits diminishing standalone regression returns at larger expert budgets: once expert-only fits become accurate, our fixed, un-tuned GBDT outcome model adds modest finite-sample variance in the intermediate regime ($n \approx 2{,}875\text{--}17{,}000$) before base DSL reconverges with expert-only estimation at very large $n$ (Section~\ref{sec:dsl-results}). Adaptive regularization or cross-validated outcome-model selection could narrow that intermediate gap, though DSL's practical regression value naturally centers on low-label settings.

%% file: sections/08_ethical_considerations.tex
\section{Ethical Considerations}
\label{sec:ethical-considerations}

\paragraph{Positive societal impacts.}
Because \textsc{Lucid} calibrates evaluators through API prompts rather than weight fine-tuning, research groups with modest annotation budgets and no GPU clusters can audit chain-of-thought systems more reliably.

\paragraph{False objectivity and deployment risks.}
Statistical calibration can create a false sense of objectivity: ratings labeled ``debiased'' may appear more trustworthy than they are, and a noisy or shifted diagnostic sample can introduce new biases \cite{chang2024survey}. In high-stakes pipelines such as reinforcement-learning trace selection or production filtering, treating calibrated proxy ratings as expert judgment risks reward hacking: downstream policies or filters can optimize the proxy metric while masking residual reasoning errors. Although we report cluster-robust uncertainty and validate downstream selection (Figure~\ref{fig:teaser}), practitioners should treat corrected scores as uncertain estimates, maintain human oversight in safety-critical settings, and re-check calibration when models or prompts change.

\paragraph{Annotation bias and representativeness.}
Because both \textsc{Lucid} and DSL calibrate toward human reference labels (Section~\ref{sec:limitations}), both prompt feedback and DSL's outcome model can inherit annotator biases in formatting or cultural context and miss problem types or difficulty levels scarce in the diagnostic sample. To mitigate these risks, practitioners can compare multiple annotator pools, model annotation noise inside the DSL pseudo-outcome, and audit labeled subsets for underrepresented categories or protected attributes in problem text.

\paragraph{Documentation of artifacts.}
Appendix~\ref{app:self-correction-details} records the PRM800K model versions, API settings, and DSL hyperparameters; full prompts and remaining benchmark scripts appear in the supplementary material. Because our benchmarks test English mathematical, scientific, and web-agent traces, we rely on the original dataset releases for linguistic and demographic scope.

\paragraph{Use of AI assistants.}
We used AI assistants for coding, data analysis, and manuscript editing. The language models evaluated in this study (e.g., GPT-5-mini, GLM-5.2) are research objects and are therefore outside the scope of this disclosure. The authors reviewed all AI-assisted output and take full responsibility for the content of this paper.

%% file: sections/09_appendix_background.tex
\section{Background}
\label{app:background}

\subsection{Setup}

We adopt the setup of \citet{de-pieuchon-etal-2025-benchmarking}. Consider a corpus of $N$ steps, each with design vector $X_i \in \R^{p+1}$ (a leading intercept $1$ plus $p$ covariates) and a true soundness label $Y_i \in \{0, 1\}$. A cheap but biased predictor produces ratings $\Yhat_i$ for all $N$ steps. Only $n \leq N$ steps in a sampled subset $\mathcal{D} \subset \{1,\dots,N\}$ receive expert labels $Y_i$ (indicated by $b_i = \1\{i \in \mathcal{D}\} \in \{0, 1\}$). The parameter of interest $\theta \in \R^{p+1}$ satisfies a moment condition:
\begin{equation}
    \E[u(Y_i, X_i; \theta)] = 0,
\end{equation}
where $u$ is the moment function (here, the logistic regression score $u(Y_i, X_i; \theta) = X_i(Y_i - \sigma(X_i^\top \theta))$, where $\sigma(z) = 1/(1+e^{-z})$ is the logistic sigmoid).

We compare five estimators.
\begin{itemize}
    \item The \textbf{reference} fit ($\thetaref$) uses expert labels $Y_i$ for all $N$ steps.
    \item \textbf{Expert-only} ($\thetahat_{\mathrm{exp}}$) fits the regression on the $n$ expert-labeled steps in $\mathcal{D}$.
    \item \textbf{Imputation} ($\thetahat_{\mathrm{imp}}$, the uncorrected plug-in fit) fits the regression on all $N$ predicted ratings $\Yhat_i$. Systematic rating errors can shift its coefficients.
    \item \textbf{PPI and DSL} ($\thetahat_{\mathrm{PPI}}, \thetahat_{\mathrm{DSL}}$) each combine predictions $\Yhat_i$ across all $N$ steps with expert labels $Y_i$ on $\mathcal{D}$.
\end{itemize}

\subsection{Prediction-Powered Inference (PPI)}

For logistic regression, PPI rectifies the estimating score using the expert labels \citep{angelopoulos2023prediction}. Let $\mathcal{D}$ be a simple random sample of $n$ of the $N$ steps. On identifiable coordinates, $\thetahat_{\mathrm{PPI}}$ solves
\begin{equation}
    \begin{aligned}
    0 ={}& \frac{1}{N}\sum_{i=1}^N
        u(\Yhat_i, X_i; \thetahat_{\mathrm{PPI}})\\
       &+ \frac{1}{n}\sum_{i\in\mathcal{D}} X_i(Y_i - \Yhat_i).
    \end{aligned}
    \label{eq:ppi-logistic}
\end{equation}
The first term evaluates the score on predicted ratings across all $N$ steps; the second adds the mean expert-minus-predicted score residual on $\mathcal{D}$. At each fixed $\theta$, the expectation over the sampling draw equals the full expert-label score $\frac{1}{N}\sum_{i=1}^N u(Y_i, X_i; \theta)$. For unequal inclusion probabilities, the rectification weights the labeled residuals by their inverse inclusion probabilities \citep{kluger2025prediction}.

\subsection{Design-based Supervised Learning (DSL)}

DSL uses a cross-fitted outcome model for $\mu(X_i, \Yhat_i) = \E[Y_i \mid X_i, \Yhat_i]$ inside the regression score \citep{egami2023dsl}.

\paragraph{Step 1: Cross-fitting.}
For each fold, fit the outcome model on labeled steps in the other folds (holding out whole problems $g$ across folds). Write $\hat{\mu}_i = \hat{\mu}(X_i, \Yhat_i)$ for its held-out prediction.

\paragraph{Step 2: Pseudo-outcome.}
Form the doubly-robust pseudo-outcome
\begin{equation}
    \Ytilde_i = \hat{\mu}_i + \frac{b_i}{\pi_i}(Y_i - \hat{\mu}_i),
    \label{eq:pseudo-outcome}
\end{equation}
where $b_i = \1\{i \in \mathcal{D}\} \in \{0, 1\}$ indicates whether step $i$ is expert-labeled, and $\pi_i = \Pr(b_i = 1 \mid X_i, \Yhat_i)$ is the step inclusion probability (equal to the constant $\pi = n/N$ under simple random sampling).

\paragraph{Step 3: Moment estimation.}
Substitute $\Ytilde_i$ for $Y_i$ in the logistic score and solve $\sum_{i=1}^N u(\Ytilde_i, X_i; \theta) = 0$ for $\thetahat_{\mathrm{DSL}}$. This substitution works because $u(Y_i, X_i; \theta)$ is affine in $Y_i$.

\paragraph{Double robustness.}
Under random labeling with known positive inclusion probabilities $\pi_i$, $\thetahat_{\mathrm{DSL}}$ remains consistent for identifiable logistic coefficients even when the outcome model is misspecified \citep[Proposition~1]{egami2023dsl}. More generally, suppose the labeling mechanism is conditionally random given the observed predictors and satisfies positivity. Under standard regularity conditions, DSL remains consistent when we replace $\pi_i$ with an estimator $\hat{\pi}_i$, provided we consistently estimate either the inclusion probabilities $\pi_i$ or the conditional outcome mean $\mu(X_i, \Yhat_i)$ \citep{bang2005doubly}.

\subsection{Outcome Model}

Standard DSL implementations often use honest forests \citep{wager2018estimation,athey2019generalized}; for our step-level regressions, we instead fit a two-fold cross-fitted gradient-boosted classifier (50 trees, max depth 3). Its held-out predictions $\hat{\mu}_i$ enter the residual correction in Equation~\eqref{eq:pseudo-outcome}.

\subsection{Evaluation Metrics}

Let $\mathcal{J}$ index the identifiable coordinates of $\theta$. For each experimental repetition we measure coefficient error by the Root Mean Squared Error,
\begin{equation}
    \RMSE(\thetahat) = \sqrt{ \frac{1}{|\mathcal{J}|}\sum_{j \in \mathcal{J}} \left(\thetahat_j - \thetaref_j\right)^2 },
\end{equation}
and report its mean and standard error across repetitions. Following \citet{de-pieuchon-etal-2025-benchmarking}, one can also normalize coordinate-wise on non-zero reference coordinates ($\thetaref_j \neq 0$) via the standardized Root Mean Squared Error,
\begin{equation}
    \sRMSE(\thetahat) = \sqrt{ \frac{1}{|\mathcal{J}|}\sum_{j \in \mathcal{J}} \E\!\left[ \left( \frac{\thetahat_j - \thetaref_j}{\thetaref_j} \right)^2 \right] },
\end{equation}
where the expectation is over experimental repetitions.

%% file: sections/10_appendix_features.tex
\section{Feature Specification}
\label{app:features}

We represent each reasoning step $i$ with $p = 9$ covariates, stacked with a leading intercept into $X_i \in \R^{p+1} = \R^{10}$ (or $p = 24$, $X_i \in \R^{25}$ when the 15 $k$-means embedding cluster indicators of Appendix~\ref{app:adaptive} are included):

\paragraph{Step type.} We encode step $i$'s category $s_i \in \{1,\dots,C\}$ ($C = 6$, with Setup as the reference category) using five one-hot dummies: Computation, Inference, Verification, Strategy, and Restatement. GPT-5-mini classifies each step into one of these six types, as described in Section~\ref{sec:step-soundness}.

\paragraph{Position.} We compute step $i$'s normalized position $\mathrm{pos}_i \in [0, 1]$ by dividing its step index by the trace step count, capturing where in the reasoning process the step occurs.

\paragraph{Step length.} We measure step length as the natural-log-transformed token count of step $i$'s text, $\mathrm{len}_i = \log(1 + \mathrm{tokens}_i)$, to capture verbosity effects on soundness. For descriptive visualization, we separately bin raw character counts in Appendix~\ref{app:bias-analysis}.

\paragraph{Difficulty.} We encode problem difficulty level with two one-hot dummies, using easy as the reference category: medium (MATH level 3) and hard (MATH levels 4--5). Because PRM800K drops the MATH \texttt{level} field, we assign every problem the default level~3 (medium). Consequently, the hard dummy is constant at zero in our experiments, and the medium dummy absorbs the intercept shift for this dataset.

\paragraph{Intercept.} We include an intercept term in all logistic regression models.

\begin{table}[h]
\centering
\small
\resizebox{\columnwidth}{!}{%
\begin{tabular}{lll}
\toprule
\textbf{Feature} & \textbf{Type} & \textbf{Dimension} \\
\midrule
(Intercept) & constant & 1 \\
Step type & 5 one-hot dummies & 5 \\
Normalized position ($\mathrm{pos}_i$) & continuous $\in [0, 1]$ & 1 \\
Log step length ($\mathrm{len}_i$) & continuous $\in \R_{\ge 0}$ & 1 \\
Difficulty & 2 one-hot dummies & 2 \\
\midrule
Total & & 10 (9 + intercept) \\
\bottomrule
\end{tabular}%
}
\caption{Feature summary.}
\label{tab:features}
\end{table}

%% file: sections/11_appendix_proof_double_robustness.tex
\section{Estimating Sound-Step Transitions}
\label{app:sound-pairs}

\paragraph{Pair inclusion probabilities and estimator.}
Let $b_{A,\ell} \in \{0,1\}$ indicate that consecutive within-trace pair $\ell \in \{1,\dots,M\}$ enters the validation sample with expert labels for both adjacent steps $(i(\ell), i(\ell)+1)$ in the same trace. Let $W_\ell$ collect the observed pair features: both assigned step types $s_{\ell,1}, s_{\ell,2} \in \{1,\dots,C\}$, both model ratings $\Yhat_{i(\ell)}, \Yhat_{i(\ell)+1}$, and any variables used for sampling. The marginal pair inclusion probability $\pi_A(W_\ell)=\Pr(b_{A,\ell}=1\mid W_\ell)$ (abbreviated $\pi_{A,\ell}$) depends on the sampling design:
\begin{itemize}
    \item \emph{Pair-level sampling:} drawing $m$ of the $M$ within-trace pairs uniformly without replacement and validating both endpoints gives $\pi_{A,\ell} = \pi_A = m/M$.
    \item \emph{Step-level sampling:} drawing $n$ of the $N$ individual steps uniformly without replacement and retaining pairs with both endpoints sampled gives $\pi_{A,\ell} = \pi_A = n(n-1)/[N(N-1)]$.
\end{itemize}
Define $A_\ell = Y_{i(\ell)} Y_{i(\ell)+1} \in \{0,1\}$, and let $\hat{\mu}_{A,\ell} = \hat{\mu}_A(W_\ell)$ estimate $\mu_A(W_\ell)=\E[A_\ell\mid W_\ell]$ via cross-fitting that holds out whole problems $g$ across folds. When the design probability $\pi_A(W_\ell)$ is known, we set $\hat{\pi}_{A,\ell} = \pi_A(W_\ell)$; otherwise, we cross-fit $\hat{\pi}_{A,\ell}$ from $W_\ell$ in the same folds. The augmented inverse-probability estimator \citep{bang2005doubly,egami2023dsl} restated from Equation~\eqref{eq:transition-estimate-main} is
\begin{align}
    \Atilde_\ell &= \hat{\mu}_{A,\ell} + \frac{b_{A,\ell}}{\hat{\pi}_{A,\ell}}(A_\ell-\hat{\mu}_{A,\ell}), \label{eq:pair-pseudo-outcome}\\
    \That_{cc'} &=
        \frac{\sum_{\ell:s_{\ell,1}=c}\1\{s_{\ell,2}=c'\}\,\Atilde_\ell}
             {\sum_{\ell:s_{\ell,1}=c}\Atilde_\ell}. \label{eq:transition-estimate}
\end{align}
Each numerator estimates the count of sound $c$-to-$c'$ pairs, and summing these numerators across destinations $c'$ yields the denominator---the estimated count of sound pairs leaving $c$. Their ratio $\That_{cc'}$ therefore targets $T_{cc'} = \lambda_{cc'}/\sum_{c''=1}^{C}\lambda_{cc''}$ in Equation~\eqref{eq:sound-transition}.

\begin{proposition}[Doubly-robust consistency of sound-step transitions]
\label{prop:sound-pairs}
Fix $c \in \{1,\dots,C\}$ with $\sum_{c'=1}^{C}\lambda_{cc'} > 0$ and suppose $b_{A,\ell} \perp A_\ell \mid W_\ell$ for every pair $\ell$. Let $h$ and $\pitilde_A$ be measurable functions satisfying $|h(W_\ell)| \le C_h < \infty$ and $\pi_A(W_\ell), \pitilde_A(W_\ell) \ge \epsilon > 0$, and define $\Atilde_\ell^{\circ} = h(W_\ell) + \frac{b_{A,\ell}}{\pitilde_A(W_\ell)}(A_\ell - h(W_\ell))$ and $Z_{c',\ell} = \1\{s_{\ell,1}=c, s_{\ell,2}=c'\}\,\Atilde_\ell^{\circ}$. Suppose that as the number of consecutive pairs $M \to \infty$: (i) $\min_{1 \le \ell \le M} \hat{\pi}_{A,\ell} \ge \epsilon$ holds with probability tending to $1$; (ii) the nuisance estimators converge in mean square, $\frac{1}{M}\sum_{\ell=1}^M (\hat{\mu}_{A,\ell} - h(W_\ell))^2 \xrightarrow{p} 0$ and $\frac{1}{M}\sum_{\ell=1}^M (\hat{\pi}_{A,\ell} - \pitilde_A(W_\ell))^2 \xrightarrow{p} 0$; and (iii) for every $c'$, the cell average satisfies the weak law $\frac{1}{M}\sum_{\ell=1}^M (Z_{c',\ell} - \E[Z_{c',\ell}]) \xrightarrow{p} 0$ under the sampling design and within-problem dependence. If either $\pitilde_A(W_\ell) = \pi_A(W_\ell)$ a.s.\ for all $\ell$ or $h(W_\ell) = \mu_A(W_\ell)$ a.s.\ for all $\ell$, then $\That_{cc'} \xrightarrow{p} T_{cc'}$ for every $c'$.
\end{proposition}

\begin{proof}
Fix $c, c' \in \{1,\dots,C\}$ and abbreviate $I_{cc',\ell} = \1\{s_{\ell,1}=c, s_{\ell,2}=c'\}$, $h_\ell = h(W_\ell)$, $\pitilde_{A,\ell} = \pitilde_A(W_\ell)$, $\pi_{A,\ell} = \pi_A(W_\ell)$, and $\mu_{A,\ell} = \mu_A(W_\ell)$.

\emph{Step 1 (Doubly-robust identification for fixed limits).}
Under $b_{A,\ell} \perp A_\ell \mid W_\ell$, $\E[b_{A,\ell}(A_\ell - h_\ell) \mid W_\ell] = \pi_{A,\ell}(\mu_{A,\ell} - h_\ell)$, so the conditional bias of $\Atilde_\ell^{\circ} = h_\ell + \frac{b_{A,\ell}}{\pitilde_{A,\ell}}(A_\ell - h_\ell)$ satisfies
\begin{align}
    \E[\Atilde_\ell^{\circ} - A_\ell \mid W_\ell]
    &= h_\ell + \frac{\pi_{A,\ell}}{\pitilde_{A,\ell}}(\mu_{A,\ell} - h_\ell) - \mu_{A,\ell} \nonumber\\
    &= \left(1 - \frac{\pi_{A,\ell}}{\pitilde_{A,\ell}}\right)(h_\ell - \mu_{A,\ell}).
    \label{eq:pair-dr-identity}
\end{align}
If either $\pitilde_A = \pi_A$ a.s.\ or $h = \mu_A$ a.s., then $\E[\Atilde_\ell^{\circ} \mid W_\ell] = \E[A_\ell \mid W_\ell]$ a.s. Because $(s_{\ell,1}, s_{\ell,2})$ is observed in $W_\ell$, $I_{cc',\ell}$ is $W_\ell$-measurable, and the tower property yields
\begin{align*}
    \E[I_{cc',\ell}\,\Atilde_\ell^{\circ}]
    &= \E\!\big[I_{cc',\ell}\,\E[\Atilde_\ell^{\circ} \mid W_\ell]\big] \\
    &= \E[I_{cc',\ell}\,A_\ell] \\
    &= \Pr(s_{\ell,1}=c, s_{\ell,2}=c', A_\ell=1).
\end{align*}
Summing over $c' \in \{1,\dots,C\}$ gives $\E[\1\{s_{\ell,1}=c\}\,\Atilde_\ell^{\circ}] = \Pr(s_{\ell,1}=c, A_\ell=1)$.

\emph{Step 2 (Nuisance estimation error).}
Adding and subtracting $\frac{b_{A,\ell}}{\hat{\pi}_{A,\ell}}(A_\ell - h_\ell)$ gives the exact decomposition
\begin{align*}
    \Atilde_\ell - \Atilde_\ell^{\circ}
    &= (\hat{\mu}_{A,\ell} - h_\ell)\!\left(1 - \frac{b_{A,\ell}}{\hat{\pi}_{A,\ell}}\right) \\
    &\quad + b_{A,\ell}(A_\ell - h_\ell)\,\frac{\pitilde_{A,\ell} - \hat{\pi}_{A,\ell}}{\hat{\pi}_{A,\ell} \pitilde_{A,\ell}}.
\end{align*}
Consider the event $\{\min_{1 \le \ell \le M} \hat{\pi}_{A,\ell} \ge \epsilon\}$, whose probability tends to $1$. On this event, $b_{A,\ell}, A_\ell \in \{0,1\}$, $|h_\ell| \le C_h$, and $\pitilde_{A,\ell} \ge \epsilon$ yield $|1 - b_{A,\ell}/\hat{\pi}_{A,\ell}| \le 1 + \epsilon^{-1}$ and $|b_{A,\ell}(A_\ell - h_\ell)/(\hat{\pi}_{A,\ell} \pitilde_{A,\ell})| \le (1 + C_h)\epsilon^{-2}$ (while $|I_{cc',\ell}| \le 1$). By the triangle inequality and Cauchy--Schwarz,
\begin{align*}
    &\left|\frac{1}{M}\sum_{\ell=1}^M I_{cc',\ell}(\Atilde_\ell - \Atilde_\ell^{\circ})\right| \\
    &\quad \le \left(1 + \frac{1}{\epsilon}\right)\!\left(\frac{1}{M}\sum_{\ell=1}^M (\hat{\mu}_{A,\ell} - h_\ell)^2\right)^{1/2} \\
    &\qquad + \frac{1 + C_h}{\epsilon^2}\!\left(\frac{1}{M}\sum_{\ell=1}^M (\hat{\pi}_{A,\ell} - \pitilde_{A,\ell})^2\right)^{1/2}
    \xrightarrow{p} 0.
\end{align*}

\emph{Step 3 (Ratio consistency).}
By Step~1, $\E[I_{cc',\ell}\,\Atilde_\ell^{\circ}] = \Pr(s_{\ell,1}=c, s_{\ell,2}=c', A_\ell=1)$ for every pair $\ell$, so condition~(iii) and the definition of $\lambda_{cc'}$ give $\frac{1}{M}\sum_{\ell=1}^M I_{cc',\ell}\,\Atilde_\ell^{\circ} \xrightarrow{p} \lambda_{cc'}$. Combining this convergence with Step~2, we obtain
\begin{align*}
    \frac{1}{M}\sum_{\ell=1}^M I_{cc',\ell}\,\Atilde_\ell
    &= \frac{1}{M}\sum_{\ell=1}^M I_{cc',\ell}\,\Atilde_\ell^{\circ} + o_p(1) \\
    &\xrightarrow{p} \lambda_{cc'}.
\end{align*}
Summing over the finite set $c' \in \{1,\dots,C\}$ yields $\frac{1}{M}\sum_{\ell=1}^M \1\{s_{\ell,1}=c\}\,\Atilde_\ell \xrightarrow{p} \sum_{c''=1}^{C}\lambda_{cc''} > 0$. Dividing the numerator and denominator of Equation~\eqref{eq:transition-estimate} by $M$ and applying the continuous mapping theorem gives $\That_{cc'} \xrightarrow{p} \lambda_{cc'}/\sum_{c''=1}^{C}\lambda_{cc''} = T_{cc'}$.
\end{proof}

\paragraph{Remarks on sampling regimes and finite samples.}
Condition~(iii) is mild because $|Z_{c',\ell}| \le C_h + (1+C_h)/\epsilon$. For instance, it holds when problems $g$ are independent, the sampling design assigns labels independently across problems, and no single problem dominates (i.e., $\sum_g M_g^2/M^2 \to 0$ for problem pair counts $M_g$). Under these conditions, the variance of the cell average stays at most $(C_h + (1+C_h)/\epsilon)^2 \sum_g M_g^2/M^2$.
When the design probability $\pi_{A,M}$ is known ($\hat{\pi}_{A,\ell} = \pi_{A,M}$) and vanishes as $M \to \infty$, we can relax the uniform lower bound $\pi_{A,M} \ge \epsilon > 0$ provided the expected number of validated pairs satisfies $M \pi_{A,M} \to \infty$. This relaxation also requires fitting $\hat{\mu}_A$ out-of-fold so that the weighted remainder vanishes in probability. Under pair-level sampling, $M \pi_{A,M} \to \infty$ requires $m \to \infty$; under uniform step-level sampling ($\pi_{A,M} = n(n-1)/[N(N-1)]$), it requires $n / \sqrt{N} \to \infty$. When $n = O(\sqrt{N})$, step-level sampling yields only $O(1)$ validated consecutive pairs in expectation and therefore necessitates pair-level sampling. In our corpus ($N=594{,}690$), sampling $n=1{,}000$ steps gives $n/\sqrt{N}\approx 1.3$ and fewer than $n^2/N\approx 1.7$ validated pairs in expectation. For $\That_{\mathrm{DSL}}$, we instead draw $m=1{,}000$ pairs uniformly without replacement---each with expert labels on both endpoints---and use the known $\pi_{A,\ell} = m/M$ ($\hat{\pi}_{A,\ell} = \pi_{A,\ell}$). Finally, because $\Atilde_\ell$ can be negative when $b_{A,\ell}=1$ and $A_\ell=0$, finite-sample ratios in Equation~\eqref{eq:transition-estimate} are not guaranteed to lie in $[0,1]$ without non-negative projection and row renormalization.

%% file: sections/12_appendix_iterative_dsl.tex
\section{Iterative DSL Outcome-Model Refinement}
\label{app:iterative-dsl}

\subsection{Method}

DSL produces two outputs: debiased coefficients $\thetahat_{\mathrm{DSL}}$ and cross-fitted outcome-model predictions $\hat{\mu}(X_i, \Yhat_i)$. These predictions are less biased than the original binary rating $\Yhat_i$ because the gradient-boosted model learns to correct systematic errors in $\Yhat_i$ using step features $X_i$. Motivated by this property, we iteratively refine continuous proxy scores $\hat{P}_i^{(t)} \in [0, 1]$ across statistical rounds $t$. Unlike \textsc{Lucid}'s prompt-based LLM re-grading, which updates $\Yhat_i^{(\tau)}$ across prompt rounds $\tau$ (Section~\ref{sec:pref-correction}), this statistical procedure runs as follows:

\begin{enumerate}
    \item Initialize $\hat{P}_i^{(0)} = \Yhat_i$ (original biased predictor).
    \item For round $t = 0, 1, 2, \ldots$:
    \begin{enumerate}
        \item Run DSL with $\hat{P}_i^{(t)}$ as the proxy predictor.
        \item Record imputation $\RMSE$ (error of $\thetahat_{\mathrm{imp}}^{(t)}$) and DSL $\RMSE$ (total coefficient error of $\thetahat_{\mathrm{DSL}}^{(t)}$).
        \item Set $\hat{P}_i^{(t+1)} = \clip_{[0,1]}\!\big(\hat{\mu}^{(t)}(X_i, \hat{P}_i^{(t)})\big)$.
    \end{enumerate}
\end{enumerate}

\paragraph{Intuition and theoretical caveat.}
Round $t = 0$ replaces the raw binary proxy $\Yhat_i$ with the cross-fitted estimate $\hat{P}_i^{(1)} = \hat{\mu}^{(0)}(X_i, \Yhat_i)$ of $\E[Y_i \mid X_i, \Yhat_i]$. By the tower property, an exact round-0 model is already a fixed point of $\E[Y_i \mid X_i, \hat{P}_i^{(1)}]$, so additional rounds $t \ge 1$ on the same labeled sample $\mathcal{D}$ cannot introduce new population information and also reuse $\mathcal{D}$ across folds (breaking strict cross-fitting independence for $t \ge 1$). We therefore treat rounds $t \ge 1$ as an exploratory diagnostic to verify that a single cross-fitted pass ($t = 0 \to 1$) already exhausts the calibration gain.

\paragraph{Two distinct benefits.}
Iterative DSL refinement serves two purposes: (1) \emph{prediction cleaning}, producing calibrated probabilities $\hat{P}_i^{(t)}$ for downstream use (e.g., as training data or trajectory scores), and (2) \emph{estimator refinement}, updating the DSL coefficient estimates $\thetahat_{\mathrm{DSL}}^{(t)}$. As shown below, these two benefits scale differently: predictor bias drops substantially (79\% at $n = 10{,}000$), while coefficient $\RMSE$ improves only modestly (3\%), because DSL already removes first-order bias in the first round.

\subsection{Overall Convergence Results}

We evaluate iterative DSL refinement at three labeling budgets: $n \in \{200, 1{,}000, 10{,}000\}$, with 4 rounds ($t \in \{0, 1, 2, 3\}$) and 10 repetitions per setting.

\begin{table}[h]
\centering\small
\begin{tabular}{lrrrr}
\toprule
$\boldsymbol{n}$ & \textbf{Round 0} & \textbf{Round 1} & \textbf{Round 2} & \textbf{Round 3} \\
\midrule
\multicolumn{5}{l}{\textit{Imputation $\RMSE$ (predictor bias)}} \\
200  & 0.786 & 0.691 & 0.694 & 0.702 \\
1{,}000  & 0.786 & 0.306 & 0.304 & 0.306 \\
10{,}000 & 0.786 & 0.172 & 0.170 & 0.168 \\
\midrule
\multicolumn{5}{l}{\textit{DSL $\RMSE$ (coefficient error)}} \\
200  & 0.712 & 0.714 & 0.716 & 0.669 \\
1{,}000  & 0.338 & 0.332 & 0.332 & 0.332 \\
10{,}000 & 0.280 & 0.290 & 0.279 & 0.272 \\
\bottomrule
\end{tabular}
\caption{Iterative DSL convergence across rounds $t = 0, \dots, 3$. Imputation $\RMSE$ measures predictor error; DSL $\RMSE$ measures total coefficient estimation error (bias and variance). Predictor bias drops 61--79\% from $t=0$ to $t=3$, while coefficient $\RMSE$ improves only 2--6\%.}
\label{tab:iterative-app}
\end{table}

\paragraph{Prediction cleaning.}
Table~\ref{tab:iterative-app} shows that iterative refinement dramatically reduces predictor bias. At $n = 10{,}000$, imputation $\RMSE$ drops from 0.786 to 0.168 (79\% reduction) over 3 rounds. At $n = 1{,}000$, the drop is 61\% (0.786 to 0.306). The largest improvement occurs between $t = 0$ and $t = 1$; subsequent rounds yield diminishing returns.

\paragraph{Coefficient estimates.}
In contrast, DSL $\RMSE$ improves by just 2--6\% across all $n$ values. DSL's pseudo-outcome construction already corrects predictor bias in the first round. Specifically, as long as the selection probability $\pi_i = \pi$ is correct, the IPW term $(b_i/\pi_i)(Y_i - \hat{\mu}_i)$ removes the bias of $\hat{\mu}_i$.

\subsection{Per-Step-Type Analysis}

To understand which reasoning types benefit most from iterative correction, we analyze accuracy improvements on the $n_{\mathrm{eval}} = 1{,}000$ evaluation set (same set used for prompt self-correction in Section~\ref{sec:iterative-prompt-results}). For each step type, we compare GPT-5-mini's original binary accuracy with the accuracy of thresholded predictions $\hat{P}_i^{(t)}$ after each round of iterative DSL refinement. Because 91.6\% of evaluation steps are sound, thresholded binary accuracy here reflects calibration alignment rather than minority-class error detection (as in Table~\ref{tab:prm-transfer}, hard thresholding at 0.5 does not improve $F_{1,-}$, so downstream pipelines should retain continuous probabilities $\hat{P}_i^{(1)}$).

\begin{table}[h]
\centering\small
\resizebox{\columnwidth}{!}{%
\begin{tabular}{lrrrr}
\toprule
\textbf{Step Type} & \textbf{Original} & \textbf{Round 1} & \textbf{Round 2} & $\boldsymbol{\Delta}$ \\
\midrule
Verification & 73.1\% & 85.1\% & 88.1\% & $+$14.9pp \\
Computation & 75.9\% & 85.2\% & 86.0\% & $+$10.1pp \\
Setup & 84.9\% & 88.2\% & 89.1\% & $+$4.2pp \\
Strategy & 82.7\% & 84.8\% & 86.4\% & $+$3.7pp \\
Inference & 77.8\% & 79.5\% & 79.5\% & $+$1.7pp \\
Restatement & 82.2\% & 83.3\% & 84.4\% & $+$2.2pp \\
\bottomrule
\end{tabular}%
}
\caption{Per-step-type accuracy across iterative DSL rounds ($\Delta = \text{Round 2} - \text{Original}$). Verification and computation benefit most, consistent with having the largest initial biases.}
\label{tab:iterative-per-type-app}
\end{table}

\paragraph{Per-type gains.}
Table~\ref{tab:iterative-per-type-app} shows that verification steps see the largest accuracy gain ($+$14.9pp by round 2), followed by computation ($+$10.1pp). On the held-out evaluation set, verification and computation also start with the largest initial biases ($-$11.9pp and $-$14.6pp, respectively). Types already well-calibrated (setup, restatement) see minimal improvement ($+$2--4pp).

\paragraph{Type-specific convergence.}
Verification accuracy continues to improve from round 1 to round 2 ($+$3.0pp additional gain), whereas computation accuracy plateaus after round 1. This contrast suggests that the outcome model requires an extra refinement round to fully adjust predictions on verification steps.

\subsection{Position Bias Correction}

GPT-5-mini systematically under-rates later steps in reasoning traces. We examine whether iterative correction addresses this position bias across quintiles (Table~\ref{tab:iterative-per-position-app}).

\begin{table}[h]
\centering\small
\resizebox{\columnwidth}{!}{%
\begin{tabular}{lrrrr}
\toprule
\textbf{Quintile} & \textbf{Original Acc} & \textbf{Round 1} & \textbf{Round 2} & $\boldsymbol{\Delta}$ \\
\midrule
0.0--0.2 & 84.2\% & 85.9\% & 86.1\% & $+$1.9pp \\
0.2--0.4 & 83.0\% & 84.1\% & 84.1\% & $+$1.1pp \\
0.4--0.6 & 78.0\% & 81.6\% & 82.4\% & $+$4.4pp \\
0.6--0.8 & 71.0\% & 78.2\% & 80.5\% & $+$9.5pp \\
0.8--1.0 & 58.8\% & 71.9\% & 75.6\% & $+$16.8pp \\
\bottomrule
\end{tabular}%
}
\caption{Position quintile accuracy across iterative rounds ($\Delta = \text{Round 2} - \text{Original Acc}$). Later quintiles benefit most from iterative correction.}
\label{tab:iterative-per-position-app}
\end{table}

\paragraph{Position bias.}
By round 2, accuracy in the final quintile (0.8--1.0) rises by $+$16.8pp---the largest positional gain---directly offsetting the rater's late-position bias ($r = -0.25$). Earlier quintiles (0.0--0.4) see minimal improvement ($+$1--2pp), as they were already well-calibrated.

\paragraph{Interaction with step type.}
Iterative DSL corrects late-position bias most strongly on computation and verification steps, which compound type and position biases, while adjusting setup and strategy steps minimally.

\subsection{Convergence Rate Analysis}

\begin{figure}[h]
    \centering
    \includegraphics[width=\columnwidth]{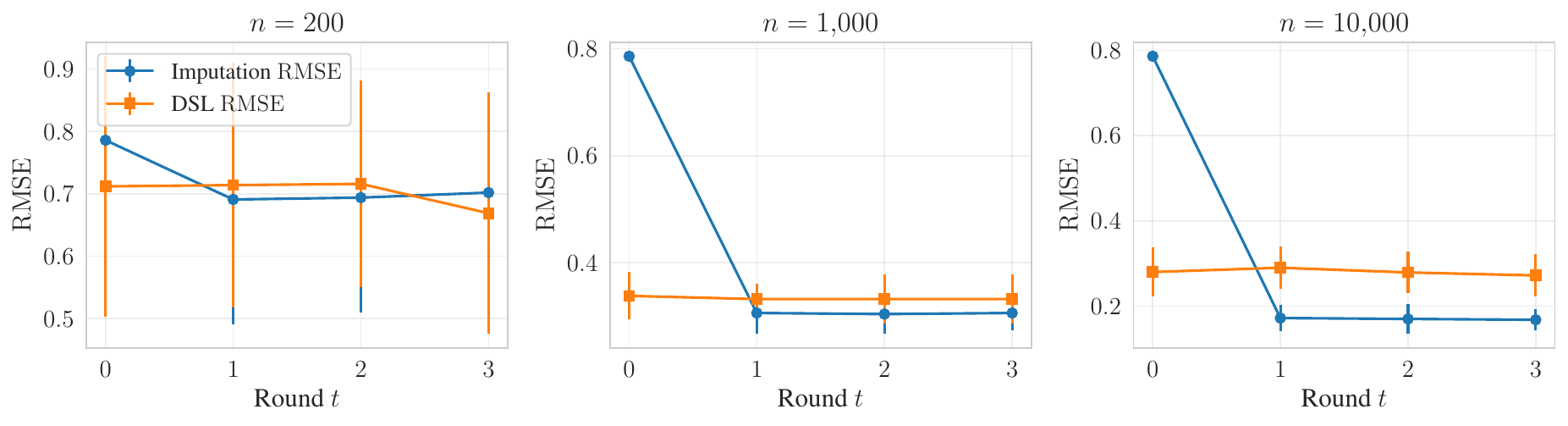}
    \caption{Iterative DSL convergence across rounds $t$ at different labeling budgets $n$. Predictor bias (imputation $\RMSE$) drops rapidly at $t=1$, then stabilizes. DSL $\RMSE$ is nearly flat from the start.}
    \label{fig:convergence-rate-app}
\end{figure}

\paragraph{Round-by-round gains.}
Figure~\ref{fig:convergence-rate-app} shows that the largest bias reduction occurs between $t = 0$ and $t = 1$. At $n = 10{,}000$, round 1 reduces imputation $\RMSE$ by 0.614 (78\% of its round-0 level, i.e., essentially the entire 0.618 total reduction). Round 2 adds only 0.002 additional reduction. This pattern holds across all $n$ values.

\paragraph{Effect of $n$.}
At $n = 10{,}000$ and $n = 1{,}000$, round 1 delivers over 99\% of the total imputation-$\RMSE$ reduction, and subsequent rounds shift $\RMSE$ by at most 0.004 (though per-type accuracy on verification still edges up through round 2). At $n = 200$, the outcome model cannot estimate reliable corrections from just 200 labeled examples. Consequently, gains remain modest and noisy, and rounds 2--3 slightly degrade $\RMSE$.

\paragraph{When to use iterative correction.}
Practitioners should use iterative refinement when downstream tasks---such as training data, process reward models, or ranking---require calibrated step predictions. In that setting, one round at $n \geq 1{,}000$ captures nearly all of the gain. When only regression coefficients are required, standard one-round DSL already removes first-order predictor bias in expectation.

%% file: sections/13_appendix_adaptive.tex
\section{Adaptive Feature Selection}
\label{app:adaptive}

\subsection{Method}

Augmenting the feature set with 15 embedding cluster indicators---obtained via $k$-means ($k{=}15$) over \texttt{all-MiniLM-L6-v2} embeddings of each step and its problem text (Appendix~\ref{app:features})---reduces DSL $\RMSE$ at $n = 200$ (0.837 $\to$ 0.653), but leaves a residual 0.065 offset at $n = 594{,}689$ ($N - 1$). Because $\hat{\mu}_i$ cancels out to $O(1/N)$ as $n \to N$ under Equation~\eqref{eq:pseudo-outcome}, this residual offset reflects adding the 15 cluster indicators to the regression design $X_i$ (altering the fitted specification relative to the $p = 9$ baseline), while also affecting residual variance. We therefore treat this comparison as requiring further theoretical analysis and evaluate an adaptive strategy that uses cluster features only below a threshold $n_{\mathrm{thresh}}$ and hand-crafted features otherwise.

\paragraph{Strategies compared.}
We consider three strategies:
\begin{itemize}
    \item \textbf{Base}: always use $p = 9$ hand-crafted features.
    \item \textbf{Full}: always use $p = 24$ features (hand-crafted + cluster).
    \item \textbf{Adaptive}: use $p = 24$ when $n < n_{\mathrm{thresh}}$, $p = 9$ otherwise.
\end{itemize}

\paragraph{Threshold selection.}
The crossover threshold $n_{\mathrm{thresh}}$ trades off potential low-budget gains from cluster features against the specification shift and extra variance observed at larger $n$. In our experiments we set $n_{\mathrm{thresh}} = 5{,}000$ post hoc from the observed crossover where cluster features begin to trail the base set. 

\paragraph{Rationale.}
At small $n$, the cross-fitted outcome model has few labeled examples to train on, so grouping steps by embedding cluster can help pool bias structure across unlabeled steps. At large $n$, hand-crafted features alone suffice and converge cleanly to zero error without altering the downstream regression specification.

\subsection{Overall Results}

We evaluate adaptive feature selection at $n_{\mathrm{thresh}} = 5{,}000$, comparing against always-base ($p = 9$) and always-full ($p = 24$) strategies. All experiments use 30 repetitions with the DSL estimator.

\begin{table}[h]
\centering\small
\resizebox{\columnwidth}{!}{%
\begin{tabular}{lrrrrr}
\toprule
$\boldsymbol{n}$ & \textbf{base ($\boldsymbol{p=9}$)} & \textbf{full ($\boldsymbol{p=24}$)} & \textbf{adaptive} & \textbf{winner} \\
\midrule
200   & 0.837 & 0.653 & 0.748 & full \\
486   & 0.449 & 0.461 & 0.482 & base \\
1{,}182  & 0.293 & 0.341 & 0.334 & base \\
2{,}875  & 0.328 & 0.313 & 0.304 & adaptive \\
6{,}993  & 0.405 & 0.335 & 0.387 & full \\
17{,}006  & 0.190 & 0.181 & 0.194 & full \\
41{,}355  & 0.070 & 0.094 & 0.069 & adaptive \\
100{,}566 & 0.033 & 0.075 & 0.035 & base \\
244{,}552 & 0.019 & 0.073 & 0.018 & adaptive \\
594{,}689 & 0.000 & 0.065 & 0.000 & base \\
\bottomrule
\end{tabular}%
}
\caption{DSL $\RMSE$ for base ($p=9$), full ($p=24$), and adaptive feature strategies. Always-base wins at 4/10 $n$-values, while always-full and adaptive each win at 3/10; no single strategy dominates across all $n$.}
\label{tab:adaptive-app}
\end{table}

\paragraph{Comparison.}
Across the 10 budgets in Table~\ref{tab:adaptive-app}, always-base wins four times, always-full wins three times, and adaptive wins three times. Because the base, full, and adaptive columns come from separate 30-repetition runs, differences between adaptive and its matching fixed strategy (full for $n < 5{,}000$; base for $n \ge 5{,}000$) reflect Monte Carlo run-to-run variation rather than distinct estimators. The effect of cluster features is non-monotonic: they help at $n = 200$, $2{,}875$, $6{,}993$, and $17{,}006$, yet trail at $n = 486$, $1{,}182$, and $n \geq 41{,}355$, with the persistent 0.065 floor at $n = 594{,}689$ reflecting the augmented $X_i$ specification.

\subsection{Cluster Characterization}

To understand what the 15 cluster features capture, we characterize each cluster by size, bias, and dominant properties.

\begin{table}[h]
\centering\small
\resizebox{\columnwidth}{!}{%
\begin{tabular}{rrrrr}
\toprule
\textbf{Cluster} & \textbf{Steps} & \textbf{Bias (pp)} & \textbf{True Rate} & \textbf{Pred Rate} \\
\midrule
12 & 77{,}392 & $-$7.1pp & 92.5\% & 85.3\% \\
8 & 59{,}447 & $-$4.6pp & 88.8\% & 84.2\% \\
3 & 53{,}528 & $-$11.6pp & 95.0\% & 83.3\% \\
10 & 47{,}687 & $-$10.7pp & 91.2\% & 80.6\% \\
7 & 45{,}835 & $-$6.1pp & 91.9\% & 85.8\% \\
9 & 42{,}280 & $-$10.3pp & 91.4\% & 81.2\% \\
1 & 38{,}709 & $-$7.4pp & 90.9\% & 83.5\% \\
0 & 36{,}555 & $-$5.8pp & 90.7\% & 84.9\% \\
6 & 34{,}738 & $-$9.1pp & 90.9\% & 81.8\% \\
13 & 34{,}286 & $-$6.8pp & 92.5\% & 85.8\% \\
5 & 33{,}559 & $-$13.4pp & 88.8\% & 75.4\% \\
11 & 31{,}497 & $-$8.9pp & 90.6\% & 81.7\% \\
4 & 28{,}893 & $-$7.4pp & 89.9\% & 82.5\% \\
2 & 20{,}795 & $-$10.5pp & 93.3\% & 82.8\% \\
14 & 9{,}489 & $-$4.1pp & 88.0\% & 83.8\% \\
\bottomrule
\end{tabular}%
}
\caption{Cluster characterization. Clusters differ substantially in step difficulty (true rates 88.0--95.0\%) and under-rating bias ($-$4.1pp to $-$13.4pp); cluster 5 shows the most severe bias.}
\label{tab:cluster-chars-app}
\end{table}

\paragraph{Cluster interpretation.}
Table~\ref{tab:cluster-chars-app} shows that the 15 embedding clusters separate steps by difficulty: clusters 3 and 13 contain mostly correct steps (95.0\% and 92.5\% true rates), while clusters 5, 8, and 14 are the hardest (true rates 88.0--88.8\%). GPT-5-mini under-rates cluster 5 most severely, predicting 75.4\% vs.\ 88.8\% true (signed bias $-$13.4pp, calibration gap 13.4pp), and under-rates cluster 14 least, predicting 83.8\% vs.\ 88.0\% (signed bias $-$4.1pp, gap 4.1pp). In the largest cluster (cluster 12), it predicts 85.3\% vs.\ 92.5\% true, leaving a 7.1pp calibration gap. No cluster exhibits over-rating: the model under-rates steps across all 15 clusters.

\paragraph{Why clusters help at small $n$ and leave an offset at large $n$.}
With few labeled examples, cluster indicators group steps that share similar bias patterns, letting the outcome model pool information across unlabeled steps in the same cluster. At large $n$, however, hand-crafted features alone already drive base DSL error to zero, whereas augmenting the regression design $X_i$ with all 15 cluster indicators introduces both collinearity and a persistent specification offset.

\subsection{Practical Recommendation}

For most deployments we recommend the base feature set ($p = 9$), which is simpler, faster, and converges cleanly to zero error as $n \to N$ without altering the downstream regression specification. Embedding cluster features showed a significant gain only at our smallest budget ($n = 200$, cutting $\RMSE$ by 22\%); future work should explore how to incorporate auxiliary cluster features into the first-stage outcome model $\hat{\mu}$ in a more informative manner. 

%% file: sections/14_appendix_rag.tex
\section{RAG-Augmented Prediction}
\label{app:rag}

\subsection{Setup}

We explore whether retrieval-augmented generation (RAG) can reduce predictor bias \emph{before} debiasing. In this approach, before the model rates each step, the pipeline retrieves similar human-labeled examples from a labeled pool and presents them as few-shot context that can potentially calibrate the model's ratings.

\paragraph{Method.} From the $N = 594{,}690$ steps, we randomly select a labeled pool of 1{,}000 steps and a query set of 9{,}000 steps. For each query step, we embed the problem text and step text using \texttt{all-MiniLM-L6-v2} (384-dimensional), retrieve the $k = 5$ most similar labeled examples via FAISS cosine similarity, and construct a few-shot prompt with their human-annotated labels. The same model (GPT-5-mini) rates the query step with this augmented prompt. We compare the RAG predictor's accuracy and bias profile against the non-RAG (zero-shot) predictor.

\subsection{Overall Results}

\paragraph{Naive RAG.}
Retrieval-augmented few-shot prompting drops GPT-5-mini's accuracy by 21.6pp, from 76.6\% (non-RAG) to 55.0\% (RAG). With RAG, the rater assigns ``correct'' only 56.5\% of the time, compared to the true rate of 91.2\% and the non-RAG rate of 82.5\%. Both prompted raters behave as near-constant classifiers that barely separate correct from incorrect steps: the non-RAG rater predicts ``correct'' $\sim$83\% of the time regardless of the true label, while RAG shifts this rate to $\sim$57\%. Neither prompted rater varies its ratings reliably with step features.

\paragraph{Bias profile.}
Table~\ref{tab:rag-bias-app} compares the logistic regression coefficients fitted on non-RAG GPT-5-mini ratings ($\thetahat_{\mathrm{imp}}$) and RAG ratings ($\thetahat_{\mathrm{RAG}}$) against the expert reference ($\thetaref$) across five representative covariates. Whereas non-RAG ratings inflate the fitted logistic regression intercept relative to $\thetaref$ (0.24 $\to$ 0.48), RAG makes the rater uniformly conservative across step features and collapses all five fitted coefficients to $\le 0.06$.

\begin{table}[h]
\centering\small
\begin{tabular}{lrrr}
\toprule
\textbf{Coefficient} & $\boldsymbol{\thetaref}$ & $\boldsymbol{\thetahat_{\mathrm{imp}}}$ & $\boldsymbol{\thetahat_{\mathrm{RAG}}}$ \\
\midrule
(Intercept)          & 0.24 & 0.48 & 0.05 \\
Step type (computation) & 0.24 & 0.48 & 0.05 \\
Normalized position  & 0.28 & 0.04 & 0.06 \\
Log step length      & 0.46 & 0.03 & 0.02 \\
Difficulty (medium)  & 0.24 & 0.48 & 0.05 \\
\bottomrule
\end{tabular}
\caption{Selected logistic regression coefficients for the expert reference ($\thetaref$), non-RAG GPT-5-mini imputation ($\thetahat_{\mathrm{imp}}$), and RAG-augmented ($\thetahat_{\mathrm{RAG}}$) predictors. RAG collapses the listed coefficients toward zero, worsening discrimination.}
\label{tab:rag-bias-app}
\end{table}

\subsection{Interventions and Root Cause Decomposition}

To diagnose why RAG degrades accuracy, we compare what the two prompts supply to GPT-5-mini and test three interventions. Specifically, we (1) restore prior steps from the same trace to the RAG prompt to match the original rating format, (2) restrict retrieval to same-problem examples with smaller $k = 3$, and (3) add embedding-derived cluster features to the DSL feature matrix (evaluated separately in Appendix~\ref{app:adaptive}). Although retrieved examples match the labeled pool's label distribution (92.6\% correct), naive RAG replaces the trace's prior steps with surface-similar steps from \emph{different} problems (e.g., sharing algebraic notation), depriving the rater of the target reasoning chain and introducing irrelevant context.

Accordingly, our prompt-level interventions suggest a two-factor breakdown of the 21.6pp drop:

\begin{enumerate}
    \item \textbf{Missing prior-steps context} (14.9pp): The original rating prompt provides all prior steps from the same trace so GPT-5-mini sees the full reasoning chain before it evaluates the current step. Restoring prior-steps context to the RAG prompt recovers 14.9pp (raising accuracy from 55.0\% to 69.9\%), making context loss the dominant factor.
    \item \textbf{Irrelevant examples} (6.7pp): Even with prior-steps context restored, RAG still trails zero-shot prompting by 6.7pp (69.9\% vs.\ 76.6\%) because retrieved examples from unrelated problems distract the rater.
\end{enumerate}

\paragraph{Why same-problem retrieval helps marginally.}
Restricting retrieval to the same problem adds only $+$0.7pp (raising accuracy from 69.9\% to 70.6\% with prior steps restored). Because the 1{,}000-step labeled pool spans only $\sim$90 problems, most queries find at most 1--2 same-problem matches, leaving at least 1--2 of the $k = 3$ examples from different problems. Overall, retrieval-augmented few-shot prompting is ineffective for step-level soundness rating in our setting.

\subsection{When Might RAG Work?}

Our results suggest RAG is unlikely to help for step-level soundness rating in its current form. However, RAG might be effective under different conditions:

\begin{itemize}
    \item \textbf{Preserved trace context with same-problem matches}: RAG avoids the 14.9pp context penalty if it retains prior steps. To overcome the remaining 6.7pp drag from unrelated examples, however, practitioners would need a much larger labeled pool (e.g., 10K+ problems) so that retrieved examples come from the same problem.
    \item \textbf{Domain-specific retrieval}: In narrower subdomains (e.g., geometry or calculus), restricting retrieval by topic might reduce irrelevant matches.
    \item \textbf{Process reward model training}: Retrieved expert labels may be more effective as supervised training data for a dedicated verifier than as in-context examples.
\end{itemize}

\paragraph{Recommendation.}
We recommend against RAG-augmented prompting for step-level soundness rating: zero-shot prompting with prior-step context outperforms every RAG variant. When a large labeled pool is available, training a dedicated process reward model is preferable.

%% file: sections/15_appendix_sensitivity.tex
\section{Self-Correction Sensitivity Analysis}
\label{app:self-correction-sensitivity}

This appendix examines the robustness of prompt-based self-correction across seven dimensions: diagnosis set size, feedback wording, profile staleness (outdated bias summaries losing accuracy under distribution shift), diagnosis granularity, expert-label noise, cross-model taxonomy agreement, and perturbation/white-noise controls. In the primary sensitivity runs, we evaluate GPT-5-mini on the same 1{,}000-step PRM800K eval set, whereas the staleness and taxonomy checks use dedicated subsets. At $n_{\mathrm{diag}}{=}1{,}000$, GPT-5-mini reaches 84.3\% in the multi-round pipeline harness, 85.6--86.2\% in the wording-ablation harness, and 85.3\% $\pm$ 0.4\% in the canonical single-round harness (headline 85.7\%; Section~\ref{sec:iterative-prompt-results}). Because 1--2pp differences across harnesses reflect prompt-wrapper formatting and pipeline configuration rather than API sampling noise alone (within-harness SD is 0.1--0.4pp), we compare variants within each harness.

\subsection{Sensitivity to Diagnosis Set Size}
\label{sec:n-diag-sensitivity}

We vary the number of expert-labeled steps used for bias diagnosis ($n_{\mathrm{diag}}$) while keeping $n_{\mathrm{eval}} = 1{,}000$ fixed.

\begin{table}[h]
\centering\small
\begin{tabular}{lrrr}
\toprule
$\boldsymbol{n_{\mathrm{diag}}}$ & \textbf{Original Acc} & \textbf{Corrected Acc} & $\boldsymbol{\Delta}$ \\
\midrule
500  & 79.0\% & 83.4\% & $+$4.4pp \\
1{,}000  & 79.0\% & 84.3\% & $+$5.3pp \\
2{,}000  & 79.0\% & 84.6\% & $+$5.6pp \\
\bottomrule
\end{tabular}
\caption{Sensitivity to diagnosis set size. More labels help, with diminishing returns.}
\label{tab:n-diag-sensitivity}
\end{table}

\paragraph{Diminishing returns.}
Increasing $n_{\mathrm{diag}}$ from 500 to 1{,}000 adds $+$0.9pp, whereas increasing $n_{\mathrm{diag}}$ from 1{,}000 to 2{,}000 adds only $+$0.3pp (Table~\ref{tab:n-diag-sensitivity}). Consequently, $n_{\mathrm{diag}} = 1{,}000$ captures most of the bias structure, and additional labels provide diminishing returns.

\paragraph{Minimum viable $n_{\mathrm{diag}}$.}
Even with only 500 labels, self-correction achieves $+$4.4pp (83\% of the $n_{\mathrm{diag}} = 1{,}000$ improvement). For practical deployment, $n_{\mathrm{diag}} = 500$ may be a reasonable cost-accuracy tradeoff.

\subsection{Sensitivity to Feedback Wording}
\label{sec:wording-sensitivity}

We test three feedback variants built from the same PRM800K diagnostic profile, each in four independent runs (Table~\ref{tab:wording-sensitivity}):

\begin{enumerate}
    \item \textbf{Detailed}: This variant presents full per-step-type biases with exact percentages (``Computation: true 94.8\%, you predict 80.0\%. Bias: $-$14.8\%.'').
    \item \textbf{Concise}: This variant issues short directives (``Please increase ratings for computation and verification steps. Be less conservative for later steps.'').
    \item \textbf{Minimal}: This variant reports overall calibration and per-type magnitudes without a full profile (``You are too conservative (predict 83.1\%, true is 91.3\%). Especially computation: you under-rate by 14.8pp.'').
\end{enumerate}

\begin{table*}[h]
\centering\small
\begin{tabular}{lrr}
\toprule
\textbf{Variant} & \textbf{Acc (4 runs)} & $\boldsymbol{\Delta}$ \\
\midrule
Detailed & 86.2\% $\pm$ 0.2 & $+$7.2pp $\pm$ 0.2 \\
Concise  & 85.6\% $\pm$ 0.1 & $+$6.6pp $\pm$ 0.1 \\
Minimal  & 85.7\% $\pm$ 0.1 & $+$6.7pp $\pm$ 0.1 \\
\midrule
No feedback & 79.0\% & --- \\
\bottomrule
\end{tabular}
\caption{Sensitivity to feedback wording, mean $\pm$ SD over four runs per variant. A paired bootstrap over evaluation steps detects no significant differences across wordings, yielding $+$0.5pp for detailed $-$ minimal (95\% CI [$-$0.6, $+$1.7]), $+$0.5pp for detailed $-$ concise ([$-$0.7, $+$1.7]), and 0.0pp for concise $-$ minimal ([$-$1.1, $+$1.1]).}
\label{tab:wording-sensitivity}
\end{table*}

\paragraph{Insensitivity to wording.}
With four runs per variant, all three diagnosis-based wordings achieve $+$6.6--7.2pp and are statistically indistinguishable. Consistent with the component ablation (Section~\ref{sec:iterative-prompt-results}), what matters is feeding the diagnosed bias back to the rater, not how it is phrased.

\subsection{Profile Staleness Under Distribution Shift}
\label{app:staleness}

Does a bias profile diagnosed on one subset remain useful on a systematically shifted subset, or does staleness---loss of calibration effectiveness as the evaluation distribution diverges from the diagnostic sample---require re-diagnosis? We run two controlled arms (Table~\ref{tab:staleness}), each comparing shifted-subset feedback against full-feedback on the \emph{same} eval steps.

\paragraph{Arm 1: position shift.}
We diagnose bias on early steps only (normalized position $\leq 0.5$; $n_{\mathrm{diag}} = 636$) and apply the resulting feedback to a late-step eval set (position $> 0.5$; $n_{\mathrm{eval}} = 1{,}000$; original accuracy 71.2\%, signed bias $-$18.4pp). Because the rater's bias grows with position, the early-step diagnosis understates late-step bias. Nevertheless, early-diagnosed feedback raises accuracy to 78.8\% and shrinks the calibration gap to 8.8pp (signed bias $-$8.8pp), versus 79.1\% and 8.5pp ($-$8.5pp) for the full-diagnosed profile---a staleness cost of only 0.3pp. Notably, \emph{both} profiles leave an $\approx$8.5pp residual gap on late steps, so a single dataset-wide prompt (not tailored by trace position) under-corrects position-concentrated bias even without shift. Adding position-specific feedback offers a promising refinement (Section~\ref{sec:limitations}).

\paragraph{Arm 2: problem holdout.}
We diagnose bias on steps from half of the diagnostic problems ($n_{\mathrm{diag}} = 489$ across 410 problems) and evaluate on $1{,}000$ steps from unseen problems disjoint from both the diagnostic and primary evaluation sets (baseline accuracy 78.0\%, signed bias $-$6.0pp). Half-problem feedback reaches 83.6\% with a near-zero $+$0.2pp signed bias (0.2pp gap), matching or slightly exceeding full-profile feedback (83.3\%, $+$1.1pp). The profile thus generalizes across the problem population within a domain. Practitioners need to re-diagnose bias---at a cost of \$0.30 per round---only when crossing domains or evaluators (Section~\ref{sec:generalization}).

\begin{table*}[h]
\centering\small
\begin{tabular}{llrrr}
\toprule
\textbf{Eval $\boldsymbol{\leftarrow}$ diagnosis} & \textbf{Acc} & \textbf{Bias ($\boldsymbol{\deltahat_{\mathcal{E}}}$)} & $\boldsymbol{F_{1,-}}$ \\
\midrule
late $\leftarrow$ original (none)        & 71.2\% & $-$18.4pp & 0.172 \\
late $\leftarrow$ early-only ($n_{\mathrm{diag}}{=}636$) & 78.8\% & $-$8.8pp  & 0.159 \\
late $\leftarrow$ full ($n_{\mathrm{diag}}{=}1{,}000$)     & 79.1\% & $-$8.5pp  & 0.161 \\
\midrule
holdout $\leftarrow$ original (none)     & 78.0\% & $-$6.0pp & 0.120 \\
holdout $\leftarrow$ half-problems ($n_{\mathrm{diag}}{=}489$) & 83.6\% & $+$0.2pp & 0.128 \\
holdout $\leftarrow$ full ($n_{\mathrm{diag}}{=}1{,}000$)      & 83.3\% & $+$1.1pp & 0.067 \\
\bottomrule
\end{tabular}
\caption{Staleness under distribution shift. Arm 1: position shift (early diagnosis $\to$ late eval); staleness costs 0.3pp vs full diagnosis. Arm 2: problem-population shift (half-problem diagnosis $\to$ held-out problems); the profile transfers.}
\label{tab:staleness}
\end{table*}

\subsection{Per-Problem vs Per-Dataset Diagnosis}
\label{sec:per-problem}

We diagnose bias on individual problems (10 steps per problem, 50 problems) and compare with per-dataset diagnosis (Table~\ref{tab:per-problem}).

\begin{table*}[h]
\centering\small
\begin{tabular}{lrrr}
\toprule
\textbf{Diagnosis} & \textbf{Original Acc} & \textbf{Corrected Acc} & $\boldsymbol{\Delta}$ \\
\midrule
Per-dataset ($n_{\mathrm{diag}} = 1{,}000$) & 79.0\% & 84.3\% & $+$5.3pp \\
Per-problem (50 problems, 10 steps each) & 79.0\% & 80.8\% & $+$1.8pp \\
\bottomrule
\end{tabular}
\caption{Per-problem diagnosis underperforms per-dataset due to noisy bias estimates.}
\label{tab:per-problem}
\end{table*}

\paragraph{Per-problem diagnosis.}
With only 10 steps per problem, the bias profile is extremely noisy. Some problems have 100\% correct steps (no bias to diagnose), while others have mixed types (unreliable per-type estimates). The resulting feedback is often inaccurate, yielding only $+$1.8pp improvement.

\paragraph{Implication.}
For practical deployment, diagnose bias on a diverse set of problems (e.g., 1{,}000 steps across many problems) rather than per-problem. If per-problem diagnosis is needed, use at least 50--100 labeled steps per problem for reliable estimates.

\subsection{Sensitivity to Expert-Label Noise}
\label{app:label-noise}

The bias diagnosis inherits any skew in the expert reference (Section~\ref{sec:limitations}), so we test how the pipeline behaves when the $n_{\mathrm{diag}} = 1{,}000$ diagnostic labels are corrupted (Table~\ref{tab:label-noise}). We perturb the diagnostic labels using 5\%/10\% uniform random flips and a structured corruption that flips 10\% of sound computation labels to unsound to simulate systematically strict annotators. We then measure the diagnosed bias profile and run the full self-correction loop with feedback from clean vs.\ 10\%-noised labels.

Label corruption shifts diagnosed magnitudes (overall bias up to $+$8.8pp) but preserves the profile's rank order (Spearman $\rho$ of per-type biases vs.\ clean: 0.94/0.83/0.83 across the three corruptions). Inside the self-correction loop, 10\% random label noise \emph{attenuates} \textsc{Lucid}'s accuracy gain ($+$5.5pp clean vs.\ $+$2.6pp noisy) and calibration-gap reduction (calibration gap $7.8\to 1.9$pp vs.\ $7.8\to 6.6$pp; signed bias $-$7.8$\to$$-$1.9pp vs.\ $-$7.8$\to$$-$6.6pp). Even so, \textsc{Lucid} still improves error detection ($F_{1,-} = 0.171$ clean and $0.214$ noisy vs.\ $0.146$ uncorrected) and keeps post-feedback accuracy above the uncorrected 79.0\% baseline. Moderate expert-label noise thus weakens \textsc{Lucid}'s gain without reversing its direction.

As a subset-size sanity check in the same harness, \textsc{Lucid} prompts GPT-5-mini with a clean $n_{\mathrm{diag}} = 1{,}000$ diagnosis and achieves 84.5\% accuracy (1.9pp calibration gap, signed bias $-$1.9pp, $F_{1,-} = 0.171$). This closely matches the multi-round pipeline's round-1 accuracy (84.3\%; Table~\ref{tab:n-diag-sensitivity}) and sits within 0.8pp of the four-run canonical full-corpus mean (85.3\%; Table~\ref{tab:iterative-prompt}), confirming that $n_{\mathrm{diag}} = 1{,}000$ labeled steps suffice without full-corpus statistics.

\begin{table*}[h]
\centering\small
\begin{tabular}{lrrrr}
\toprule
\textbf{Diagnosis} & \textbf{Acc} & \textbf{Bias (pp)} & $\boldsymbol{\rho}$ & \textbf{Max shift} \\
\midrule
No correction           & 79.0\% & $-$7.8 & ---  & --- \\
Clean ($n_{\mathrm{diag}}{=}1{,}000$)   & 84.5\% & $-$1.9 & 1.00 & --- \\
5\% random flips        & ---    & ---    & 0.94 & 9.0pp \\
10\% random flips       & 81.6\% & $-$6.6 & 0.83 & 10.6pp \\
Structured comp.\ 10\%  & ---    & ---    & 0.83 & 9.3pp \\
\bottomrule
\end{tabular}
\caption{Expert-label noise sensitivity ($n_{\mathrm{eval}} = 1{,}000$). Bias (pp) reports signed evaluation bias $\deltahat_{\mathcal{E}}$; $\rho$ measures the Spearman correlation of per-type diagnosed biases vs.\ the clean profile; and max shift records the largest per-type bias change under the corruption. Loop accuracy shown for clean and 10\% random-flip diagnosis.}
\label{tab:label-noise}
\end{table*}

\subsection{Taxonomy Agreement: Cross-Model Proxy}
\label{app:taxonomy-agreement}

As preliminary evidence for taxonomy validity, we re-classified 200 steps (stratified across the six GPT-5-mini-assigned types) with GLM-5.2, a separate model family not used to construct the taxonomy. Overall, the two models agree moderately (Cohen's $\kappa = 0.42$; 52\% raw agreement) and unevenly across categories (Table~\ref{tab:taxonomy-agreement}). They agree closely on the largest classes---computation (91\%) and strategy (94\%)---but they rarely agree on restatement (27\%) or verification (12\%). On those two categories, GLM-5.2 typically assigns computation or inference instead. The two models repeat this pattern on a second serving stack (Azure) both when we disable ($\kappa = 0.44$) and when we enable ($\kappa = 0.40$) extended reasoning. Across those two settings, GLM-5.2 assigns ``verification'' to only 5 and 4 of the 200 steps, respectively (Table~\ref{tab:taxonomy-agreement}). Because neither the serving stack nor extended reasoning alters agreement, we disable extended reasoning in all remaining GLM-5.2 runs to reduce cost.

To test whether this category disagreement changes our bias conclusions, we perform two checks. First, when we recompute GPT-5-mini's diagnosed bias under GLM-5.2's own type labels across all three runs, GPT-5-mini still under-rates computation ($-$20.0/$-$15.5/$-$19.6pp) and inference ($-$14.3/$-$16.2/$-$12.5pp). However, because GLM-5.2 produces $\le 5$ verification assignments, we have too few cases to cross-check that category. Second, when we randomly reassign 30\% of step types (Section~\ref{sec:dsl-results}), every category retains its bias sign, confirming that our conclusions hold even when borderline steps shift categories.

Without human step-type annotations, comparing two LLMs cannot establish either model's classification accuracy; human inter-annotator validation of these 200 steps remains future work.

\begin{table*}[h]
\centering\small
\begin{tabular}{lrrrr}
\toprule
\textbf{GPT-5-mini type} & \textbf{Steps} & \textbf{Token-plan} & \textbf{Azure, no-think} & \textbf{Azure, think} \\
\midrule
Computation   & 34 & 91.2\% & 100\%  & 91.2\% \\
Strategy      & 33 & 93.9\% & 93.9\% & 93.9\% \\
Setup         & 34 & 47.1\% & 32.4\% & 35.3\% \\
Inference     & 33 & 39.4\% & 54.5\% & 39.4\% \\
Restatement   & 33 & 27.3\% & 27.3\% & 27.3\% \\
Verification  & 33 & 12.1\% & 12.1\% & 12.1\% \\
\midrule
\textbf{All}  & \textbf{200} & \textbf{52.0\% }($\kappa = 0.42$) & \textbf{53.5\% }($\kappa = 0.44$) & \textbf{50.0\% }($\kappa = 0.40$) \\
\bottomrule
\end{tabular}
\caption{Per-type agreement between GPT-5-mini (taxonomy assigner) and GLM-5.2 (independent re-classifier) on the same 200 stratified steps, across two serving stacks and both thinking settings.}
\label{tab:taxonomy-agreement}
\end{table*}

\subsection{Perturbation and White-Noise Controls}
\label{app:noise-controls}

Table~\ref{tab:noise-controls} summarizes two additional diagnostic controls referenced in Section~\ref{sec:dsl-results}. First, we randomly reassign 30\% of step-type labels across 50 trials to test robustness to taxonomy noise. Second, we replace GPT-5-mini's structured errors with 10--30\% uniform random prediction flips to verify that \textsc{Lucid} does not overfit unstructured noise.

\begin{table*}[h]
\centering\footnotesize
\resizebox{\textwidth}{!}{%
\begin{tabular}{llrr}
\toprule
\textbf{Perturbation Control} & \textbf{Target Metric} & \textbf{Unperturbed} & \textbf{Perturbed} \\
\midrule
30\% random step-type reassignment (50 trials) & Diagnosed computation bias & $-$14.6pp & $-$13.3 $\pm$ 1.2pp \\
30\% random step-type reassignment (50 trials) & Diagnosed verification bias & $-$11.9pp & $-$10.1 $\pm$ 3.3pp \\
10--30\% white-noise prediction flips & Post-feedback eval accuracy & 80.4\% (baseline) & 78.6--81.2\% ($\Delta \le +0.8$pp) \\
\bottomrule
\end{tabular}%
}
\caption{Diagnostic perturbation and white-noise controls. When we randomly reassign 30\% of step-type labels across 50 trials, the diagnosed bias sign holds for every step category. In contrast, when we replace structured rater bias with 10--30\% uniform random prediction flips, self-correction produces no meaningful gain (versus $+$6.7pp on structured GPT-5-mini bias).}
\label{tab:noise-controls}
\end{table*}

\subsection{Practical Recommendations}

Based on our sensitivity analysis, we recommend that practitioners diagnose bias per dataset rather than per problem, using $n_{\mathrm{diag}} = 1{,}000$ labeled steps for best results ($n_{\mathrm{diag}} = 500$ captures 83\% of the benefit) and $n_{\mathrm{eval}} = 1{,}000$ evaluation steps for reliable improvement estimates. Practitioners can adopt any diagnosis-based feedback wording---component and wording ablations show that format is immaterial---and need to re-diagnose only when they switch domains or evaluators (Appendix~\ref{app:staleness}).

%% file: sections/16_appendix_bias_analysis.tex
\section{Extended Bias Analysis}
\label{app:bias-analysis}

\subsection{Position Quintile Analysis}

GPT-5-mini's bias varies dramatically with step position in the reasoning trace. We divide all $N = 594{,}690$ steps into 5 quintiles by normalized position and compute bias statistics.

\begin{table}[h]
\centering\small
\resizebox{\columnwidth}{!}{%
\begin{tabular}{lrrrr}
\toprule
\textbf{Quintile} & \textbf{Steps} & \textbf{True Rate} & \textbf{Pred Rate} & \textbf{Bias (pp)} \\
\midrule
0.0--0.2 & 169{,}941 & 89.7\% & 92.4\% & $+$2.7pp \\
0.2--0.4 & 134{,}252 & 92.3\% & 88.3\% & $-$4.0pp \\
0.4--0.6 & 127{,}571 & 92.3\% & 82.0\% & $-$10.4pp \\
0.6--0.8 & 114{,}709 & 91.2\% & 74.1\% & $-$17.2pp \\
0.8--1.0 & 48{,}217 & 92.1\% & 59.9\% & $-$32.2pp \\
\bottomrule
\end{tabular}%
}
\caption{Prediction bias by position quintile. Later steps are severely under-rated.}
\label{tab:position-quintile-app}
\end{table}

\begin{figure}[h]
    \centering
    \includegraphics[width=\columnwidth]{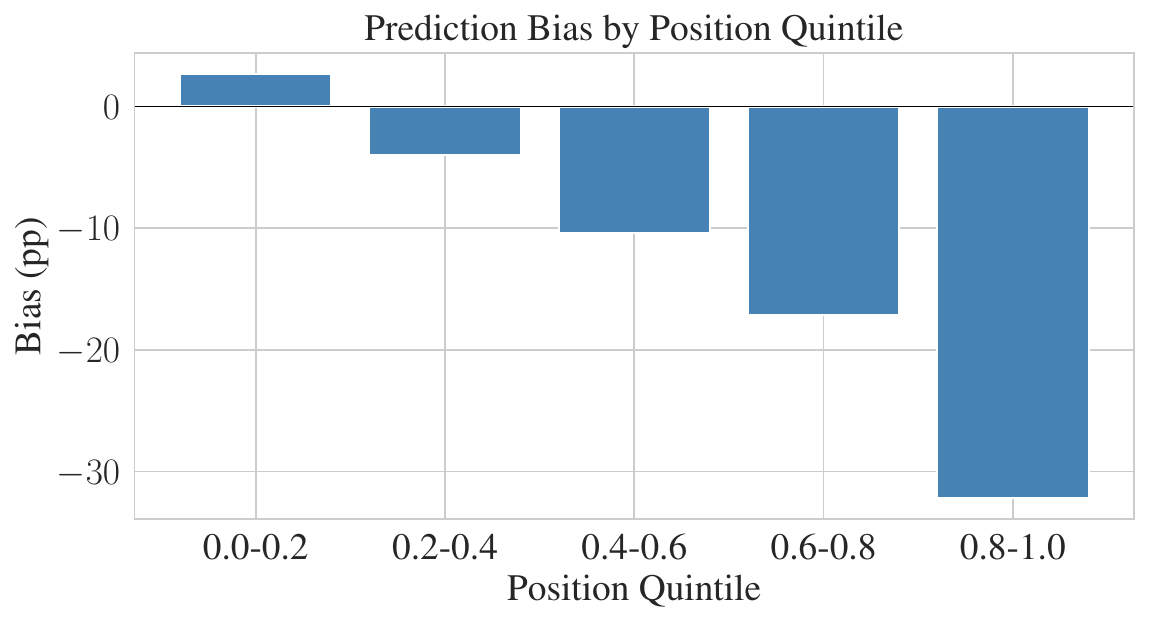}
    \caption{Prediction bias by position quintile. Later steps are systematically under-rated.}
    \label{fig:position-quintile-app}
\end{figure}

\paragraph{Position trend.}
Table~\ref{tab:position-quintile-app} shows a clear monotonic pattern: under-rating deepens monotonically from $+$2.7pp (first quintile, slight over-rating) to $-$32.2pp (last quintile, severe under-rating). In the final quintile, the model predicts ``correct'' on only 59.9\% of steps despite a 92.1\% true rate. Figure~\ref{fig:position-quintile-app} visualizes this pattern.

\paragraph{Why later steps are under-rated.}
Several factors may contribute:
\begin{itemize}
    \item \textbf{Context accumulation}: Later steps depend on more prior context; even though true soundness stays near 92\%, the rater's skepticism and false-negative errors accumulate across longer chains
    \item \textbf{Answer proximity}: The model may confuse step soundness with final answer correctness
    \item \textbf{Length correlation}: Later steps tend to be longer (more complex), which also correlates with under-rating
\end{itemize}

\subsection{Step Length Bias}

Step length also correlates with bias; whereas the regression feature $\mathrm{len}_i$ in $X_i$ uses log token count (Appendix~\ref{app:features}), here we bin steps by raw character count for direct interpretability (Table~\ref{tab:length-bias-app}).

\begin{table*}[h]
\centering\small
\begin{tabular}{lrrrr}
\toprule
\textbf{Length Bin} & \textbf{Steps} & \textbf{True Rate} & \textbf{Pred Rate} & \textbf{Bias (pp)} \\
\midrule
$<$50 chars & 93{,}250 & 90.4\% & 80.3\% & $-$10.1pp \\
50--100 chars & 253{,}448 & 91.7\% & 85.0\% & $-$6.7pp \\
100--200 chars & 229{,}984 & 91.3\% & 83.7\% & $-$7.6pp \\
200--500 chars & 17{,}820 & 92.1\% & 62.2\% & $-$29.9pp \\
500--1{,}000 chars & 158 & 84.2\% & 54.4\% & $-$29.7pp \\
$>$1{,}000 chars & 30 & 80.0\% & 50.0\% & $-$30.0pp \\
\bottomrule
\end{tabular}
\caption{Prediction bias by step length. Long steps ($>$200 chars) are severely under-rated.}
\label{tab:length-bias-app}
\end{table*}

\begin{figure}[h]
    \centering
    \includegraphics[width=\columnwidth]{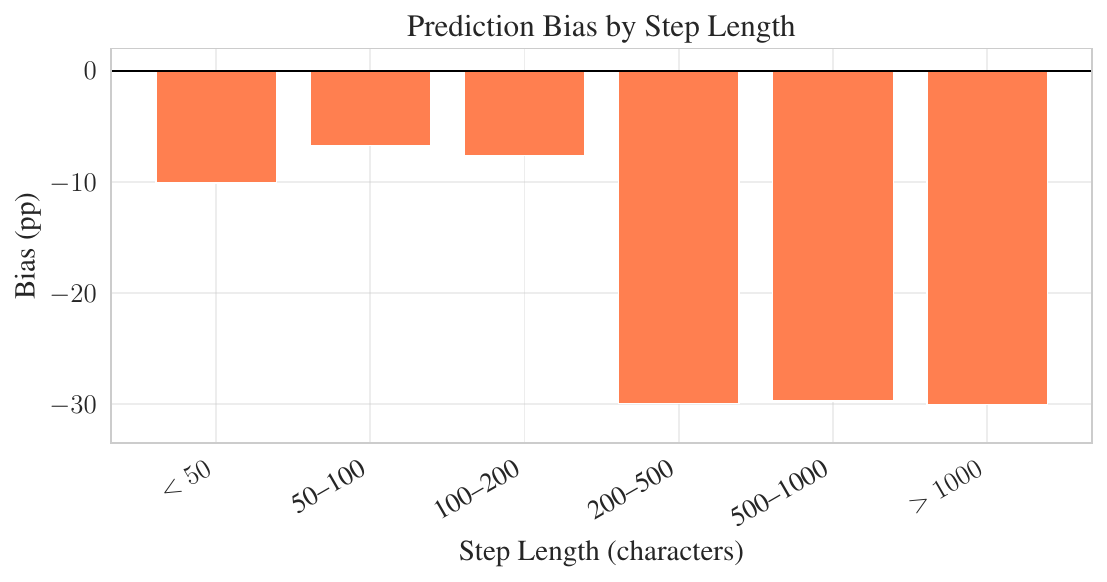}
    \caption{Prediction bias by step length. Long steps are systematically under-rated.}
    \label{fig:length-bias-app}
\end{figure}

\paragraph{Length trend.}
GPT-5-mini modestly under-rates steps up to 200 characters ($-$6.7pp to $-$10.1pp) but severely under-rates steps longer than 200 characters ($-$29.9pp to $-$30.0pp; Figure~\ref{fig:length-bias-app}). This gap widens on long steps both because they tend to appear later in a trace---compounding position bias---and because GPT-5-mini penalizes step complexity.

\paragraph{Interaction with step type.}
The length bias is strongest for computation steps (which tend to be longest) and weakest for setup/strategy steps (which tend to be shortest).

\subsection{Interaction Effects}

\begin{figure}[h]
    \centering
    \includegraphics[width=\columnwidth]{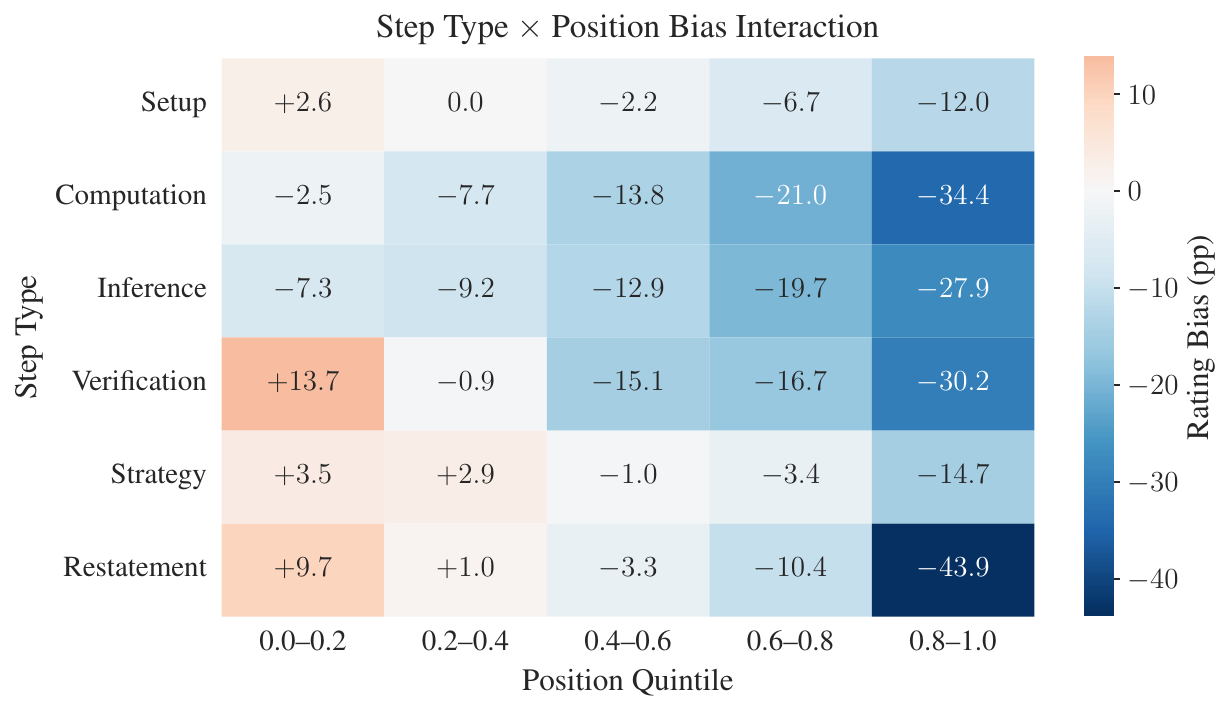}
    \caption{Step type $\times$ position bias interaction. Darker blue indicates more severe under-rating (negative bias), whereas warm orange/peach indicates over-rating (positive bias). Computation, verification, and restatement steps in late positions show the strongest under-rating.}
    \label{fig:interaction}
\end{figure}

Bias is not additive across dimensions; Table~\ref{tab:interaction-app} and Figure~\ref{fig:interaction} break it down by step type and position.

\begin{table*}[h]
\centering\small
\begin{tabular}{lrrrrr}
\toprule
\textbf{Type} & \textbf{0.0--0.2} & \textbf{0.2--0.4} & \textbf{0.4--0.6} & \textbf{0.6--0.8} & \textbf{0.8--1.0} \\
\midrule
Setup & $+$2.6pp & $-$0.0pp & $-$2.2pp & $-$6.7pp & $-$12.0pp \\
Computation & $-$2.5pp & $-$7.7pp & $-$13.8pp & $-$21.0pp & $-$34.4pp \\
Inference & $-$7.3pp & $-$9.2pp & $-$12.9pp & $-$19.7pp & $-$27.9pp \\
Verification & $+$13.7pp & $-$0.9pp & $-$15.1pp & $-$16.7pp & $-$30.2pp \\
Strategy & $+$3.5pp & $+$2.9pp & $-$1.0pp & $-$3.4pp & $-$14.7pp \\
Restatement & $+$9.7pp & $+$1.0pp & $-$3.3pp & $-$10.4pp & $-$43.9pp \\
\bottomrule
\end{tabular}
\caption{Step type $\times$ position bias (pp). Restatement and computation in late positions are most severely under-rated.}
\label{tab:interaction-app}
\end{table*}

\paragraph{Restatement in the final quintile.}
Restatement steps in the final quintile (0.8--1.0) show the largest bias ($-$43.9pp). Because these steps often summarize the final answer, the model may penalize them for repetition or for depending on earlier steps.

\paragraph{Computation.}
GPT-5-mini under-rates computation steps at every position, with bias deepening from $-$2.5pp (first quintile) to $-$34.4pp (last quintile); only restatement (53.6pp) and verification (43.9pp) show steeper drops across quintiles.

\paragraph{Verification.}
GPT-5-mini actually \emph{over-rates} verification steps in the first quintile ($+$13.7pp) before under-rating them in later quintiles. Early verification steps (checking setup) are easy and the rater grades them leniently; late verification steps (checking complex results) are harder and the rater under-rates them.

\subsection{Summary of Bias Patterns}

\begin{itemize}
    \item \textbf{Position}: GPT-5-mini shifts monotonically from $+$2.7pp (early) to $-$32.2pp (late), creating a strong position bias.
    \item \textbf{Length}: GPT-5-mini shows a threshold effect at $\sim$200 chars and under-rates long steps by $\sim$30pp.
    \item \textbf{Step type}: GPT-5-mini is most biased on verification and computation steps and least biased on setup and strategy steps.
    \item \textbf{Interaction}: Restatement $\times$ late position is worst ($-$43.9pp)
    \item \textbf{Combined}: Prompting GPT-5-mini with calibration feedback disproportionately corrects these high-bias cells (e.g., $+$25.6pp in the final position quintile).
\end{itemize}

\begin{table}[h]
\centering\small
\begin{tabular}{lr}
\toprule
\textbf{Domain / step type} & \textbf{Diagnosed bias (pp)} \\
\midrule
Mind2Web overall & $-$84.2 \\
\quad action execution & $-$82.6 \\
\quad termination & $-$98.0 \\
\midrule
ProofWriter overall & $-$4.2 \\
\quad derivation & $-$33.3 \\
\quad rule & $-$11.8 \\
\bottomrule
\end{tabular}
\caption{Diagnosed signed bias profiles ($\deltahat_{\mathcal{D}}$, $\deltahat_{\mathcal{D},c}$) on the labeled diagnostic sets $\mathcal{D}$ (the input to \textsc{Lucid}'s feedback) for the two boundary domains. Table~\ref{tab:multidomain} reports the corresponding eval-set calibration gaps ($|\deltahat_{\mathcal{E}}|$: 89.2pp Mind2Web, 4.8pp ProofWriter).}
\label{tab:boundary-profiles}
\end{table}

%% file: sections/17_appendix_transition_matrix.tex
\section{Transition Matrix Comparison}
\label{app:transition-matrix}

Across all three panels of Figure~\ref{fig:transition-heatmap}, we use the same GPT-5-mini-assigned step types. We form the expert reference $\Tref$ (panel~a) from pairs that experts rated sound, and we form the uncorrected matrix $\That_{\mathrm{uncorr}}$ (panel~b) from pairs that GPT-5-mini rated sound. For the DSL matrix $\That_{\mathrm{DSL}}$ (panel~c), we apply Equation~\eqref{eq:transition-estimate-main} to $m=1{,}000$ validated pairs. We compare each estimate against $\Tref$ using $\|\That-\Tref\|_F/\|\Tref\|_F$.

\begin{figure*}[h]
    \centering
    \includegraphics[width=\columnwidth]{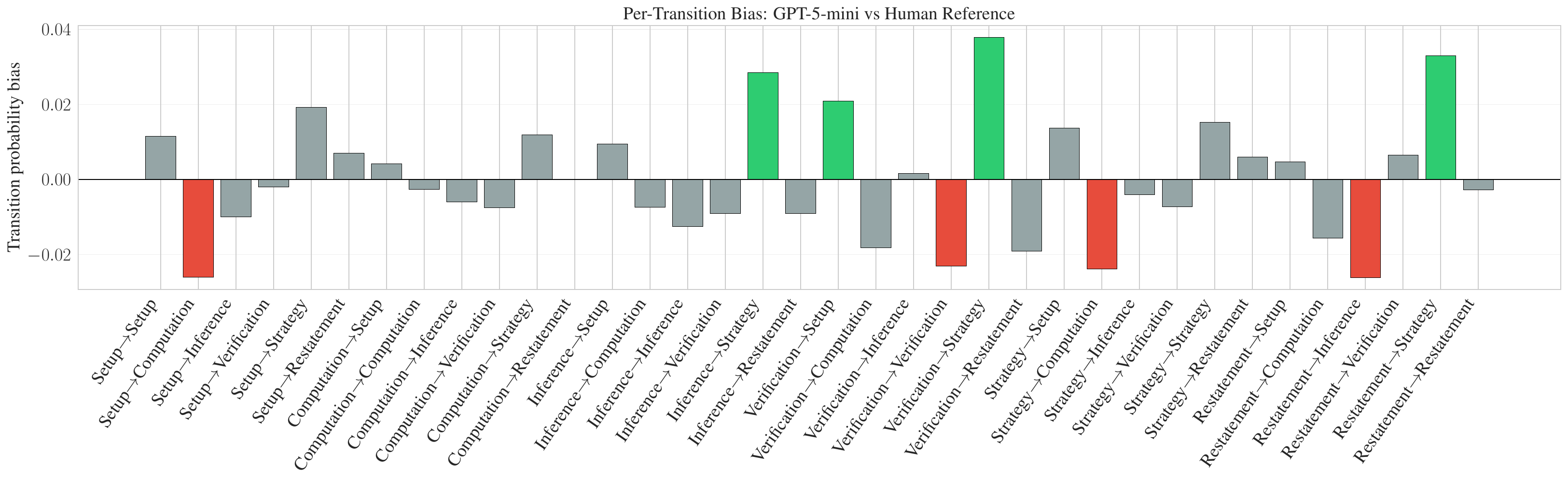}
    \caption{Per-transition bias: $\That_{\mathrm{uncorr}} - \Tref$. Red bars mark underestimates below $-0.02$, green bars mark overestimates above $+0.02$, and gray bars lie within $\pm 0.02$. Conditioning on biased ratings most strongly underestimates transitions such as Restatement $\to$ Inference, Setup $\to$ Computation, Strategy $\to$ Computation, and Verification $\to$ Verification (down to $-0.03$) while overestimating transitions into Strategy (up to $+0.04$).}
    \label{fig:transition-bias-app}
\end{figure*}

\paragraph{Per-transition bias.}
Figure~\ref{fig:transition-bias-app} shows the per-transition bias $\That_{\mathrm{uncorr}} - \Tref$ (highlighting shifts exceeding $\pm 0.02$ in red or green). GPT-5-mini grades computation, inference, and verification strictly while grading strategy and setup more leniently. Consequently, $\That_{\mathrm{uncorr}}$ most strongly underestimates Restatement $\to$ Inference ($-0.03$), Setup $\to$ Computation ($-0.03$), Strategy $\to$ Computation ($-0.02$), and Verification $\to$ Verification ($-0.02$). It also produces smaller negative shifts on other transitions, such as Verification $\to$ Computation ($-0.02$, gray bar). Conversely, $\That_{\mathrm{uncorr}}$ inflates transitions into Strategy---most notably Verification $\to$ Strategy ($+0.04$), Restatement $\to$ Strategy ($+0.03$), and Inference $\to$ Strategy ($+0.03$)---as well as Verification $\to$ Setup ($+0.02$).

%% file: sections/18_appendix_self_correction.tex
\section{Self-Correction Detailed Results}
\label{app:self-correction-details}

\paragraph{Reproducibility.}
We generated GPT-5-mini ratings via the Azure OpenAI API (model version \texttt{gpt-5.4-mini}, version 2026-03-17; temperature 0, top\_p 1.0), and we classified step types with GPT-5-mini using those same Azure settings. For cross-model transfer, we generated GLM-5.2 ratings via BytePlus ModelArk (model version \texttt{GLM-5.2}, version 2026-06-17; temperature 0). We provide all prompts in the supplementary material and summarize full experimental metadata in Table~\ref{tab:reproducibility}.

\begin{table*}[h]
\centering\footnotesize
\begin{tabular}{lll}
\toprule
\textbf{Component} & \textbf{Parameter} & \textbf{Value} \\
\midrule
\multirow{3}{*}{GPT-5-mini ratings} & Model version & \texttt{gpt-5.4-mini (version 2026-03-17)} \\
 & Temperature / top\_p & 0 / 1.0 \\
 & API & Azure OpenAI \\
\midrule
\multirow{2}{*}{GLM-5.2 ratings} & Model version & \texttt{glm-5.2} \\
 & Temperature & 0 \\
\midrule
\multirow{4}{*}{DSL (scalar)} & Cross-fitting model & GradientBoostingClassifier \\
 & Trees / max depth & 50 / 3 \\
 & Cross-fitting folds & 2 \\
 & Random seed & 42 \\
\midrule
\multirow{3}{*}{DSL (transition)} & Cross-fitting model & GradientBoostingClassifier \\
 & Trees / max depth & 30 / 3 \\
 & Cross-fitting folds & 2 \\
\midrule
\multirow{2}{*}{PPI} & Optimizer & L-BFGS-B \\
 & Loss & Logistic \\
\midrule
\multirow{4}{*}{Self-correction} & Diagnostic profile & Corpus profile (subset-verified $n_{\mathrm{diag}} = 500\text{--}2{,}000$) \\
 & Eval set size & $n_{\mathrm{eval}} = 1{,}000$ \\
 & Rounds & 1 (canonical) \\
 & Random seed & 42 \\
\midrule
\multirow{3}{*}{Vary-expert} & Repetitions & 30 \\
 & $n$ values & 200, 486, 1{,}182, 2{,}875, \ldots, 594{,}689 \\
 & Random seed & 42 \\
\midrule
Dataset & PRM800K & $N = 594{,}690$ steps \\
\bottomrule
\end{tabular}
\caption{Reproducibility checklist: full experimental metadata for all components.}
\label{tab:reproducibility}
\end{table*}

\begin{table*}[h]
\centering
\small
\begin{tabular}{lll}
\toprule
\textbf{Type} & \textbf{Description} & \textbf{Example} \\
\midrule
Setup & Define variables, set up equations & ``Let $x$ be the number of\ldots'' \\
Computation & Arithmetic, algebraic manipulation & ``$2x + 3 = 7$, so $x = 2$'' \\
Inference & Logical deduction from prior steps & ``Since $A = 90^{\circ}$, $C = 45^{\circ}$'' \\
Verification & Checking an intermediate result & ``Verify: $2^2 + 2^2 = 8 \neq 9$'' \\
Strategy & Choosing approach, planning & ``We'll use induction'' \\
Restatement & Rephrasing without new content & ``Thus the answer is 42.'' \\
\bottomrule
\end{tabular}
\caption{Step type taxonomy for mathematical reasoning.}
\label{tab:taxonomy}
\end{table*}

\begin{table*}[h]
\centering\small
\begin{tabular}{lrrrr}
\toprule
\textbf{Method} & \textbf{Setup Cost} & \textbf{Per-Pass Cost} & \textbf{Accuracy} & \textbf{Adaptable} \\
\midrule
Self-correction (1 round) & \$500 & \$0.30 & 85.7\% & Yes \\
Expert-only ($n{=}1{,}000$) & \$500 & --- & 80.1\% & No \\
Expert-only ($n{=}1{,}500$) & \$750 & --- & 84.0\% & No \\
Full expert annotation & \$300{,}000 & --- & 100\% & No \\
Fine-tuned LR/MiniLM ($n{=}1{,}000$) & \$500 + \$50 & --- & 91.6\% & Retrain \\
Fine-tuned Qwen3.5-LoRA ($n{=}1{,}000$) & \$500 + \$50+GPU & --- & 81.4\%\text{--}86.1\% & Retrain \\
\bottomrule
\end{tabular}
\caption{Cost comparison of debiasing methods. All $n{=}1{,}000$ methods share the same \$500 diagnostic expert-annotation budget, and \textsc{Lucid} adds only \$0.30 in API calls (\$500.30 total). At that total, \textsc{Lucid} costs one-third less than expert-only annotation at matched accuracy (\$750 at $n{=}1{,}500$) and $600\times$ less than full annotation (\$300{,}000). Moreover, \textsc{Lucid} incurs \$0.30 in marginal cost per re-rating pass and requires no GPU retraining.}
\label{tab:cost}
\end{table*}

\begin{table}[h]
\centering\small
\resizebox{\columnwidth}{!}{%
\begin{tabular}{lccccc}
\toprule
$\boldsymbol{K}$ & \textbf{Original} & \textbf{Expert} & \textbf{DSL} & $\boldsymbol{\Delta}_{\textbf{DSL-Expert}}$ & $\boldsymbol{\Delta}_{\textbf{DSL-Orig}}$ \\
\midrule
1 & 15.2\% & 27.5\% & 28.9\% & $+$1.4pp & $+$13.7pp \\
2 & 35.6\% & 51.5\% & 55.1\% & $+$3.6pp & $+$19.5pp \\
3 & 53.6\% & 68.8\% & 71.6\% & $+$2.8pp & $+$18.0pp \\
4 & 67.2\% & 80.0\% & 81.6\% & $+$1.6pp & $+$14.5pp \\
5 & 76.0\% & 87.0\% & 87.9\% & $+$0.9pp & $+$11.9pp \\
8 & 88.8\% & 95.4\% & 95.2\% & $-$0.2pp & $+$6.4pp \\
10 & 93.0\% & 97.4\% & 97.2\% & $-$0.2pp & $+$4.2pp \\
\bottomrule
\end{tabular}%
}
\caption{Full Best-of-$K$ trajectory selection accuracy on 7{,}348 MATH problems (deltas computed prior to one-decimal rounding). Continuous probabilities from DSL's cross-fitted outcome model match or slightly edge out unweighted binary Expert step-label averages at $K{=}1$ through $K{=}5$ (via tie-breaking among all-sound traces; Section~\ref{sec:iterative-prompt-results}), while Expert step labels slightly edge ahead at $K{=}8$ and $K{=}10$.}
\label{tab:trajectory-filtering-full}
\end{table}

\begin{table}[h]
\centering\small
\resizebox{\columnwidth}{!}{%
\begin{tabular}{lcccc}
\toprule
\textbf{PRM Training Labels} & \textbf{Binary Acc} & $\boldsymbol{F_{1,-}}$ & \textbf{ECE} $\boldsymbol{\downarrow}$ & \textbf{Top-1 Sel.} $\boldsymbol{\uparrow}$ \\
\midrule
Original GPT-5-mini & 91.6\% & 0.00 & 0.084 & 17.0\% \\
DSL outcome model ($\hat{\mu}_i$, $n{=}1{,}000$) & 91.6\% & 0.00 & 0.037 & 18.8\% \\
Expert / oracle (all $N$) & 91.6\% & 0.00 & 0.015 & --- \\
\bottomrule
\end{tabular}%
}
\caption{Downstream process reward model (PRM) transfer under the $p{=}9$ step-feature representation. Hard binary thresholding at 0.5 collapses all three models to the majority class, whereas continuous probabilities from a verifier trained on DSL outcome-model probabilities cut expected calibration error (ECE) by 56\% and lift top-1 trace selection by $+$1.8pp over a verifier trained on raw GPT-5-mini labels.}
\label{tab:prm-transfer}
\end{table}

\begin{table*}[h]
\centering\footnotesize
\begin{tabular}{lrrrl}
\toprule
\textbf{Run} & \textbf{Orig.\ Acc} & \textbf{Corr.\ Acc} & $\boldsymbol{\Delta}$ & \textbf{Notes} \\
\midrule
Canonical ($n_{\mathrm{eval}}{=}1{,}000$) & 79.0\% & 85.7\% & $+$6.7pp & Headline single-round pass \\
Run 2 ($n_{\mathrm{eval}}{=}1{,}000$) & 79.0\% & 85.1\% & $+$6.1pp & Independent API pass \\
Run 3 ($n_{\mathrm{eval}}{=}1{,}000$) & 79.0\% & 85.5\% & $+$6.5pp & Independent API pass \\
Cross-model GPT ($n_{\mathrm{eval}}{=}1{,}000$) & 79.0\% & 84.9\% & $+$5.9pp & Independent seed \\
\midrule
\textbf{Mean (4 runs)} & \textbf{79.0\%} & \textbf{85.3\%} & $\mathbf{+}$\textbf{6.3pp} & \textbf{SD 0.4pp} \\
\midrule
Replication ($n_{\mathrm{eval}}{=}5{,}000$) & --- & --- & $+$6.2pp & 95\% CI [4.7, 7.8]; verif.\ $+$14.3, comp.\ $+$10.0pp \\
\bottomrule
\end{tabular}
\caption{Single-round prompt self-correction across four independent 1{,}000-step runs (mean 85.3\%, SD 0.4pp) and an independent 5{,}000-step replication disjoint from both the diagnostic and primary evaluation sets (per-type bias--gain $r = 0.54$ across the six types). A multi-round run on the 1{,}000-step set converges at round 1 (84.3\%$\to$85.3\%, $+$1.0pp), with round 2 adding no further gain.}
\label{tab:iterative-prompt}
\end{table*}

\begin{table*}[h]
\centering\footnotesize
\begin{tabular}{lrrrrrr}
\toprule
\textbf{Method} & \textbf{Accuracy} & $\boldsymbol{\Delta}$ & $\boldsymbol{F_{1,-}}$ & \textbf{Recall}$\boldsymbol{_{-}}$ & $\boldsymbol{F_{1,+}}$ & \textbf{Balanced} \\
\midrule
Original GPT-5-mini & 79.0\% & --- & 0.146 & 0.214 & 0.880 & 52.9\% \\
Self-correction & 85.7\% & $+$6.7pp & 0.192 & 0.202 & 0.921 & 56.0\% \\
Global threshold shift & 84.6\% & $+$5.6pp & 0.083 & 0.083 & 0.916 & 50.0\% \\
Post-hoc type flips & 86.7\% & $+$7.7pp & 0.076 & 0.065 & 0.928 & 50.3\% \\
Post-hoc type$\times$pos flips & 85.8\% & $+$6.8pp & 0.075 & 0.069 & 0.923 & 49.9\% \\
Platt scaling & 91.6\% & $+$12.6pp & 0.00 & 0.00 & 0.956 & 50.0\% \\
Always predict 1 & 91.6\% & $+$12.6pp & 0.00 & 0.00 & 0.956 & 50.0\% \\
Fine-tuned LR & 91.6\% & $+$12.6pp & 0.00 & 0.00 & 0.956 & 50.0\% \\
Fine-tuned MiniLM & 91.6\% & $+$12.6pp & 0.00 & 0.00 & 0.956 & 50.0\% \\
Qwen3.5-0.8B (LoRA) & 81.4\% & $+$2.4pp & 0.238 & 0.345 & 0.894 & 62.0\% \\
Qwen3.5-2B (LoRA) & 66.2\% & $-$12.8pp & 0.228 & 0.595 & 0.784 & 66.3\% \\
Qwen3.5-4B (LoRA) & 86.1\% & $+$7.1pp & 0.206 & 0.214 & 0.924 & 58.7\% \\
Qwen3.5-9B (LoRA) & 82.9\% & $+$3.9pp & 0.182 & 0.226 & 0.905 & 56.2\% \\
\bottomrule
\end{tabular}
\caption{Baseline comparison on PRM800K ($n_{\mathrm{eval}} = 1{,}000$, 84 unsound steps). All methods use the same $n_{\mathrm{diag}} = 1{,}000$ labeled diagnostic steps ($n = n_{\mathrm{diag}}$). $F_{1,-}$ and Recall$_{-}$ treat \emph{unsound} steps ($Y_i = 0$) as the target class, whereas $F_{1,+}$ measures the $F_1$ score on \emph{sound} steps ($Y_i = 1$). Self-correction cuts false alarms from 144 to 76 while maintaining unsound-step recall, whereas accuracy-maximizing baselines (Platt, always-1, fine-tuned LR/MiniLM) detect no unsound steps. Global threshold shift adjusts the overall predicted soundness rate to match the true rate without reading text; post-hoc flip baselines randomly flip binary ratings within each step type (or type$\times$position cell) to match the diagnosed rate deltas (mean over 20 draws; SD 0.3--0.4pp), closing the gap without LLM re-rating but halving $F_{1,-}$ (Section~\ref{sec:results}). The Qwen3.5-LoRA verifiers are the only baselines that do \emph{not} maximize accuracy: we train them with class-weighted loss and select checkpoints by $F_{1,-}$ (unsound-step detection) on a held-out validation set. We report all metrics on the same held-out 1{,}000-step evaluation set. Even so, these four models detect only a fraction of the unsound steps ($F_{1,-}$ 0.18--0.24) and remain below the majority-class accuracy ceiling. Two of the four (0.8B, 2B) exceed self-correction's $F_{1,-}$ (0.238, 0.228 vs.\ 0.192), but they achieve substantially lower accuracy. Balanced accuracy = $(\mathrm{Recall}_{+} + \mathrm{Recall}_{-})/2$ accounts for class imbalance.}
\label{tab:baselines}
\end{table*}

\begin{table}[h]
\centering
\small
\begin{tabular}{lccc}
\toprule
\textbf{Method} & \textbf{Naive SE} & \textbf{Cluster SE} & \textbf{Ratio} \\
\midrule
Expert-only & 0.032 & 0.104 & 3.3$\times$ \\
PPI & 0.041 & 0.142 & 3.5$\times$ \\
DSL & 0.038 & 0.115 & 3.0$\times$ \\
\bottomrule
\end{tabular}
\caption{Step-level regression standard errors at $n=1{,}000$ under independent-step and problem-clustered calculations.}
\label{tab:clustering}
\end{table}

\begin{table*}[h]
\centering\small
\begin{tabular}{lrrrr}
\toprule
\textbf{Step Type} & \textbf{Steps} & \textbf{Original Acc} & \textbf{Corrected Acc} & $\boldsymbol{\Delta}$ \\
\midrule
Computation & 357 & 75.9\% & 86.3\% & $+$10.4pp \\
Verification & 67 & 73.1\% & 82.1\% & $+$9.0pp \\
Restatement & 90 & 82.2\% & 87.8\% & $+$5.6pp \\
Setup & 119 & 84.9\% & 89.9\% & $+$5.0pp \\
Inference & 176 & 77.8\% & 82.4\% & $+$4.5pp \\
Strategy & 191 & 82.7\% & 85.3\% & $+$2.6pp \\
\bottomrule
\end{tabular}
\caption{Per-step-type accuracy after single-round correction ($n_{\mathrm{eval}} = 1{,}000$; initial bias--gain $r = 0.79$, $p = 0.06$ across the six types). The share of initial errors corrected follows the same ordering, led by computation (59.3\%) over strategy (39.4\%) and restatement (37.5\%). It rises from 30.0\% in the first position quintile (0.0--0.2) to 65.2\% in the final quintile (0.8--1.0), while computation bias across quintiles on this set is $0.0$, $-9.1$, $-11.6$, $-15.8$, and $-22.6$pp.}
\label{tab:iterative-prompt-per-type}
\end{table*}

\begin{table}[h]
\centering\small
\resizebox{\columnwidth}{!}{%
\begin{tabular}{lrrr}
\toprule
\textbf{Feedback} & \textbf{Acc} & $\boldsymbol{\Delta}$ & \textbf{Share} \\
\midrule
No feedback (baseline) & 79.0\% & --- & --- \\
Overall only           & 85.3\% $\pm$ 0.4 & $+$6.3pp & 124\% \\
Type only              & 84.4\% $\pm$ 0.5 & $+$5.4pp & 106\% \\
Position only          & 81.3\% $\pm$ 0.3 & $+$2.3pp & 45\% \\
Full feedback          & 84.1\% $\pm$ 0.3 & $+$5.1pp & 100\% \\
\midrule
Canonical full feedback$^\dagger$ & 85.7\% & $+$6.7pp & 131\% \\
\bottomrule
\end{tabular}%
}
\caption{Ablation: accuracy improvement by feedback component, mean $\pm$ SD over four independent runs per condition on the same 1{,}000-step eval set. Paired bootstrap comparisons over the shared eval steps find no significant differences among the informative variants (full $-$ overall $-$1.2pp, 95\% CI [$-$2.5, $+$0.1]; full $-$ type $-$0.3pp, [$-$1.4, $+$0.9]; type $-$ overall $-$0.9pp, [$-$2.1, $+$0.3]). $^\dagger$This row reports the canonical single-round run (headline result, 4-run mean 85.3\% $\pm$ 0.4\%). This run differs by 1.2--1.6pp from the ablation harness's full-feedback condition (84.1\% $\pm$ 0.3\%). Differing prompt wrappers and pipeline harnesses across experiments---rather than API sampling noise alone---drive that offset (Appendix~\ref{app:self-correction-sensitivity}).}
\label{tab:ablation-app}
\end{table}

\subsection{Class-Balanced-Subset Ablation}
\label{app:balanced-ablation}
To isolate the effect of class imbalance, we evaluate GPT-5-mini before and after \textsc{Lucid} feedback on a class-balanced subset (84 unsound $+$ 84 randomly sampled sound steps). Self-correction moves accuracy only from 57.1\% to 58.3\% and $F_{1,-}$ from 0.333 to 0.327. This result confirms that the 91.6\% base rate drives the raw-accuracy jump on full PRM800K and that prompt feedback shifts the rater's threshold without improving class-balanced discrimination.

%% file: sections/19_appendix_additional_figures.tex
\section{Additional Figures and Tables}
\label{app:additional-figures}

\begin{table*}[h]
\centering\small
\resizebox{\textwidth}{!}{%
\begin{tabular}{lcccccc}
\toprule
\textbf{Setting} & $\boldsymbol{F_{1,-}}$ \textbf{Orig$\boldsymbol{\to}$Corr} & $\boldsymbol{\Delta F_{1,-}}$ & \textbf{Rel.} $\boldsymbol{F_{1,-}}$ & \textbf{Gap} $\boldsymbol{|\deltahat_{\mathcal{E}}|}$ \textbf{Orig$\boldsymbol{\to}$Corr} & $\boldsymbol{\Delta |\deltahat_{\mathcal{E}}|}$ & \textbf{Gap Ratio} \\
\midrule
PRM800K (GPT-5-mini) & 0.146 $\to$ 0.192 & $+$0.046 & $+$32\% & 7.8pp $\to$ 0.9pp & $-$6.9pp & $8.7\times$ \\
PRM800K (GLM-5.2$^\dagger$) & 0.143 $\to$ 0.138 & $-$0.005 & $-$3\% & 1.4pp $\to$ 2.3pp & $+$0.9pp & $0.6\times$ \\
ProcessBench & 0.523 $\to$ 0.556 & $+$0.033 & $+$6\% & 13.0pp $\to$ 4.2pp & $-$8.8pp & $3.1\times$ \\
\bottomrule
\end{tabular}%
}
\caption{Calibration gap ($|\deltahat_{\mathcal{E}}|$) and unsound-step detection ($F_{1,-}$) metrics, separating additive changes ($\Delta = \text{Corr} - \text{Orig}$) from relative $F_{1,-}$ change and multiplicative gap reduction ($\text{Orig}/\text{Corr}$). Self-correction improves both calibration and $F_{1,-}$ when the rater starts conservative. $^\dagger$GLM-5.2 uses feedback generated from GPT-5-mini's bias profile; cross-model transfer improves accuracy on an already well-calibrated model without lifting $F_{1,-}$.}
\label{tab:calibration-headline}
\end{table*}

\begin{figure}[!htbp]
    \centering
    \includegraphics[width=\columnwidth]{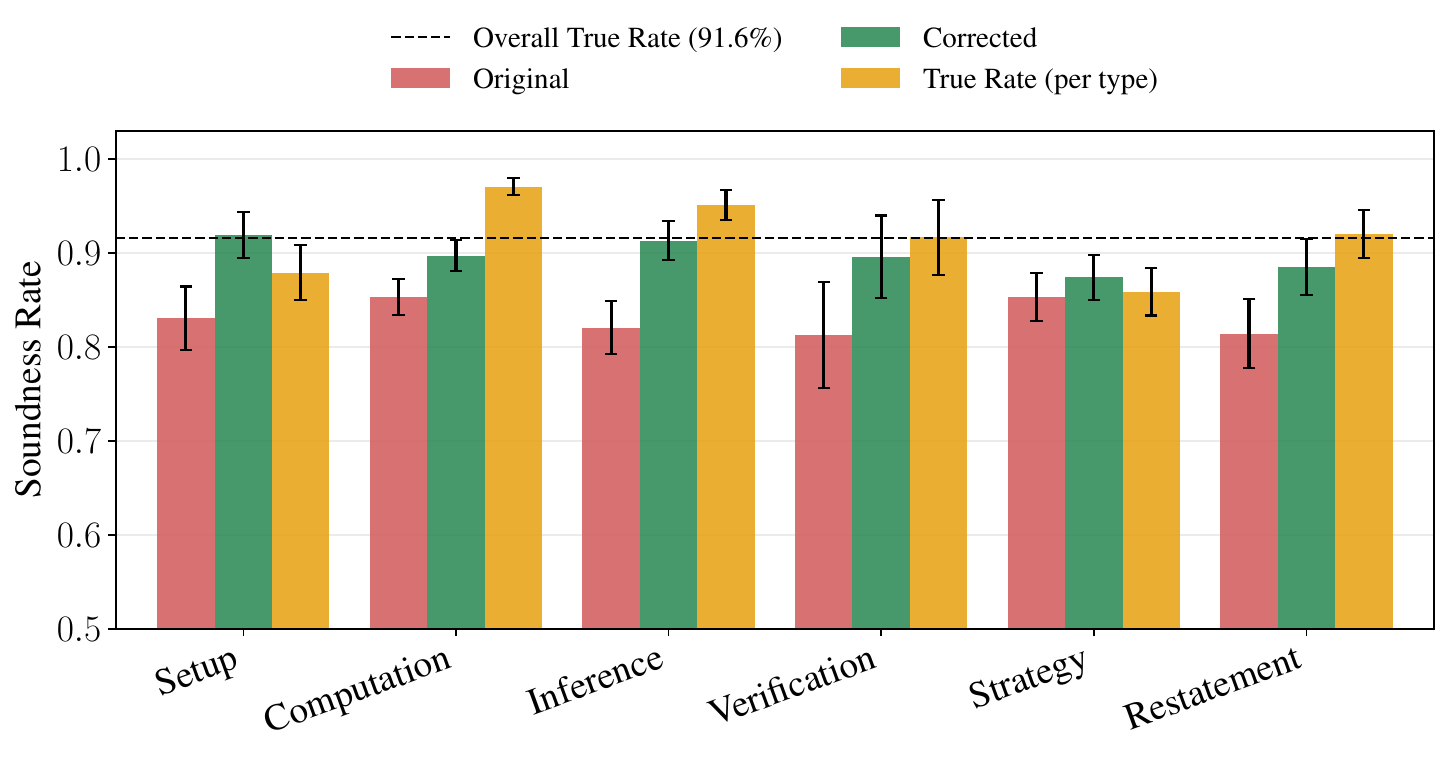}
    \caption{Calibration by step type (with $\pm 1\,\mathrm{SE}$ error bars). Predicted (Original, Corrected) vs.\ true soundness rates before and after one round of \textsc{Lucid} feedback. The dashed line marks the overall true soundness rate (91.6\%).}
    \label{fig:calibration-by-type}
\end{figure}

\begin{table}[h]
\centering
\resizebox{\columnwidth}{!}{%
\begin{tabular}{c l c c c}
\toprule
\textbf{Step} & \textbf{Type} & \textbf{GPT} & \textbf{After} & \textbf{Human} \\
\midrule
6 & Strategy & \checkmark & \checkmark & \checkmark \\
7 & Setup & \checkmark & \checkmark & \checkmark \\
\textbf{8} & \textbf{Computation} & \textcolor{red!70!black}{$\times$} & \textcolor{green!50!black}{\checkmark} & \checkmark \\
9 & Computation & \checkmark & \checkmark & \checkmark \\
15 & Computation & \checkmark & \checkmark & \checkmark \\
\bottomrule
\end{tabular}%
}
\caption{Qualitative example: \textit{``Given $f(x) = x^2 - 2x + 5$ and $g(x) = x+3$, find $f(g(5)) - g(f(5))$.''}
Step~8 (``$g(f(x)) = f(x) + 3$'') is a computation step at position~0.44 that GPT-5-mini incorrectly rates as unsound ($\times$); feedback flips it to sound (\checkmark), matching the human label.}
\label{tab:qualitative-example}
\end{table}

\begin{table*}[h]
\centering\footnotesize
\begin{tabular}{lcccccc}
\toprule
 & \multicolumn{2}{c}{\textbf{$\mathbf{RMSE}$ (mean $\boldsymbol{\pm}$ SE)}} & \multicolumn{2}{c}{\textbf{DSL vs Expert}} & \multicolumn{2}{c}{\textbf{DSL vs PPI}} \\
\cmidrule(lr){2-3} \cmidrule(lr){4-5} \cmidrule(lr){6-7}
$\boldsymbol{n}$ & \textbf{DSL} & \textbf{Expert-only} & \textbf{$\boldsymbol{t}$-test $\boldsymbol{p}$} & \textbf{Wilcoxon $\boldsymbol{p}$} & \textbf{$\boldsymbol{t}$-test $\boldsymbol{p}$} & \textbf{Wilcoxon $\boldsymbol{p}$} \\
\midrule
200 & $0.84{\pm}0.08$ & $6.09{\pm}0.99$ & $<0.001^{***}$ & $<0.001^{***}$ & $<0.001^{***}$ & $<0.001^{***}$ \\
486 & $0.45{\pm}0.03$ & $0.83{\pm}0.24$ & $0.83$ & $0.002^{**}$ & $<0.001^{***}$ & $<0.001^{***}$ \\
1{,}182 & $0.29{\pm}0.03$ & $0.31{\pm}0.02$ & $>0.99$ & $>0.99$ & $0.30$ & $<0.001^{***}$ \\
2{,}875 & $0.33{\pm}0.02$ & $0.24{\pm}0.02$ & $<0.001^{***}$ & $<0.001^{***}$ & $0.55$ & $>0.99$ \\
\bottomrule
\end{tabular}
\caption{Statistical comparison of debiasing methods (30 repetitions, paired tests, Bonferroni-corrected for 8 comparisons). $^{***}p < 0.001$, $^{**}p < 0.01$. DSL significantly outperforms both baselines at $n = 200$, outperforms Expert-only on the rank test at $n = 486$ (Wilcoxon $p = 0.002$) and matches it at $n = 1{,}182$, while Expert-only achieves modestly lower $\RMSE$ at $n = 2{,}875$ under our fixed GBDT specification (before both curves reconverge at very large $n$; Figure~\ref{fig:vary-expert}). Against PPI, DSL is significantly better on both tests at $n \in \{200, 486\}$ and on the Wilcoxon test at $n = 1{,}182$ ($p < 0.001$; paired $t$-test $p = 0.30$ due to high PPI variance), with no significant difference at $n = 2{,}875$.}
\label{tab:stat-tests}
\end{table*}

\begin{figure}[!htbp]
    \centering
    \includegraphics[width=\linewidth]{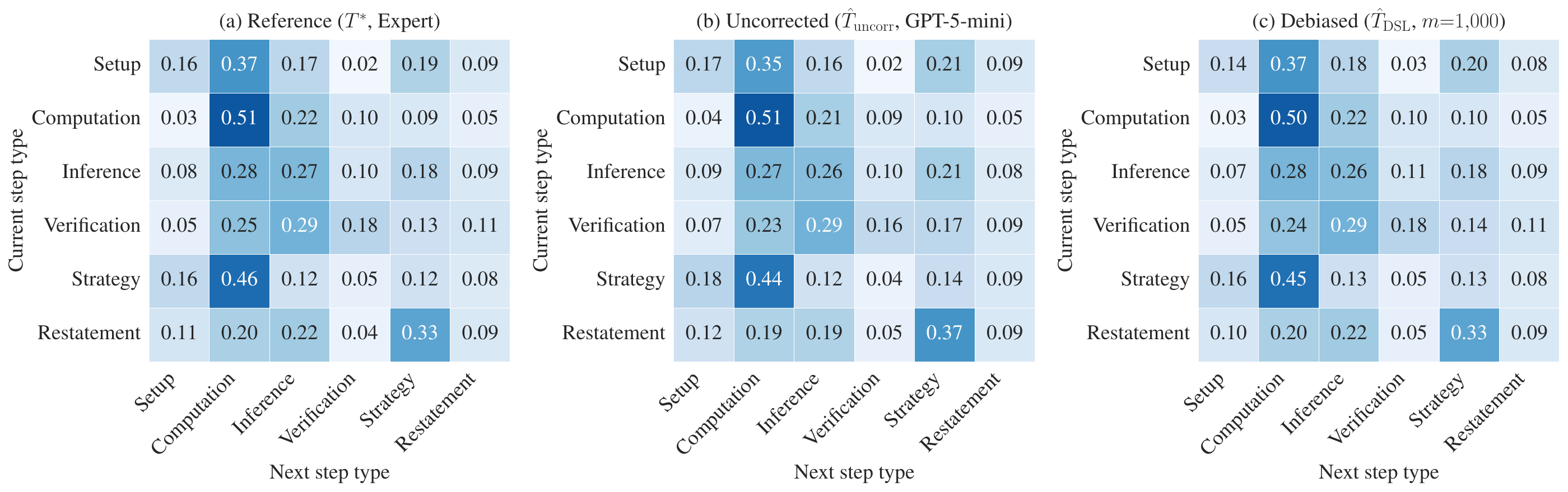}
    \caption{Transitions among the $C=6$ step types. Panel (a) shows the expert reference $\Tref$ on pairs judged sound by experts, and panel (b) shows the uncorrected matrix $\That_{\mathrm{uncorr}}$ on pairs rated sound by GPT-5-mini. Panel (c) shows the DSL estimate $\That_{\mathrm{DSL}}$ of Equation~\eqref{eq:transition-estimate-main} from $m=1{,}000$ validated pairs. Relative Frobenius errors $\|\That - \Tref\|_F/\|\Tref\|_F$ against (a) are 0.083 for (b) and 0.031 for (c) (Appendix~\ref{app:transition-matrix}).}
    \label{fig:transition-heatmap}
\end{figure}

\begin{table}[h]
\centering\small
\resizebox{\columnwidth}{!}{%
\begin{tabular}{lll}
\toprule
\textbf{Method} & \textbf{Description} & \textbf{Code} \\
\midrule
Reference & All-$N$ expert labels & IRLS \\
Expert-only & $n$ expert labels only & IRLS \\
Imputation & All-$N$ predicted labels & IRLS \\
PPI & Gradient rectification & L-BFGS-B \\
DSL & Cross-fitted GBDT + pseudo-outcome & IRLS \\
\bottomrule
\end{tabular}%
}
\caption{Estimators compared for step-level logistic regression (IRLS: Iteratively Reweighted Least Squares; GBDT: Gradient-Boosted Decision Trees).}
\label{tab:methods}
\end{table}

\subsection{Cross-Model Results}
\label{app:cross-model-details}

Tables~\ref{tab:bias-profile-cross-model} and~\ref{tab:cross-model-sc} compare the diagnosed step-rating bias profile and single-round self-correction gain across GPT-5-mini and GLM-5.2 on the shared 1{,}000-step evaluation set.

\begin{table}[!htbp]
\centering\small
\begin{tabular}{lrr}
\toprule
\textbf{Bias Dimension} & \textbf{GPT-5-mini} & \textbf{GLM-5.2} \\
\midrule
Overall (\% pred $-$ \% true) & $-$7.8pp & $-$1.4pp \\
Computation bias & $-$14.6pp & $-$2.0pp \\
Verification bias & $-$11.9pp & $-$6.0pp \\
Position correlation & $-0.25$ & $-0.12$ \\
Length correlation & $-0.03$ & $-0.08$ \\
\bottomrule
\end{tabular}
\caption{Bias profile comparison across models, computed on the shared 1{,}000-step evaluation set for both models.}
\label{tab:bias-profile-cross-model}
\end{table}

\begin{table}[!htbp]
\centering\small
\begin{tabular}{lrrrr}
\toprule
\textbf{Model} & \textbf{Original} & \textbf{Corrected} & $\boldsymbol{\Delta}$ \\
\midrule
GPT-5-mini & 79.0\% & 84.9\% & $+$5.9pp \\
GLM-5.2 & 84.4\% & 87.5\% & $+$3.1pp \\
\bottomrule
\end{tabular}
\caption{Self-correction gain across model families after a single correction round, using identical \textsc{Lucid}-diagnosed feedback (computed from GPT-5-mini's ratings). Both models benefit from the same feedback. Here, GPT-5-mini scores 84.9\% ($+$5.9pp) on a separate API pass from the 85.7\% headline pass. This variation falls within the 0.4pp run-to-run SD across four passes (Table~\ref{tab:iterative-prompt}; Section~\ref{sec:limitations}).}
\label{tab:cross-model-sc}
\end{table}

\subsection{ProcessBench Results}
\label{app:processbench-details}

\begin{table}[!htbp]
\centering\small
\resizebox{\columnwidth}{!}{%
\begin{tabular}{lrrrr}
\toprule
\textbf{Split} & \textbf{Steps} & \textbf{Original} & \textbf{Corrected} & $\boldsymbol{\Delta}$ \\
\midrule
GSM8K & 1{,}568 & 94.6\% & 92.4\% & $-$2.2pp \\
MATH & 4{,}366 & 85.9\% & 89.7\% & $+$3.8pp \\
OlympiadBench & 5{,}822 & 82.0\% & 85.7\% & $+$3.7pp \\
OmniMath & 4{,}792 & 72.3\% & 83.0\% & $+$10.7pp \\
\midrule
Total & 16{,}548 & 81.4\% & 86.6\% & $+$5.2pp \\
\bottomrule
\end{tabular}%
}
\caption{Self-correction on ProcessBench. Improvement on 3 of 4 splits; gains scale with initial bias severity.}
\label{tab:processbench}
\end{table}

\begin{figure}[h]
    \centering
    \includegraphics[width=\columnwidth]{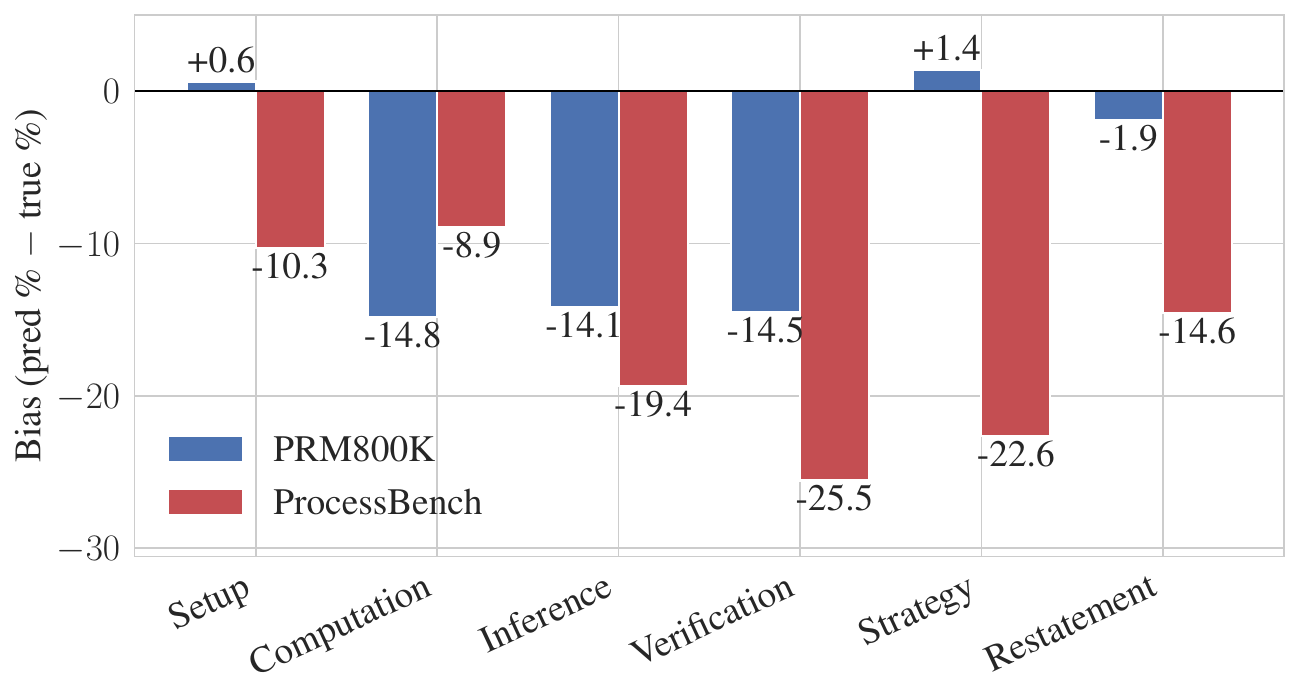}
    \caption{Per-step-type bias ($\text{pred}\% - \text{true}\%$) of GPT-5-mini on PRM800K (diagnostic profile) vs.\ ProcessBench (grouped bar chart). Under-rating occurs across all six categories on ProcessBench and across computation, inference, verification, and restatement on PRM800K, with larger gaps on ProcessBench for every type except computation.}
    \label{fig:bias-profile-comparison}
\end{figure}

\begin{table}[h]
\centering\small
\resizebox{\columnwidth}{!}{%
\begin{tabular}{lrr}
\toprule
\textbf{Method} & \textbf{Accuracy} & \textbf{95\% CI} \\
\midrule
Zero-shot & 82.8\% & [80.3, 85.0] \\
Few-shot ($k=1$) & 83.7\% & [81.3, 85.9] \\
Few-shot ($k=3$) & 83.7\% & [81.3, 85.9] \\
Few-shot ($k=5$) & 83.6\% & [81.1, 85.8] \\
Few-shot ($k=10$) & 84.0\% & [81.6, 86.1] \\
\textsc{Lucid} feedback (self-correction) & 87.9\% & [85.7, 89.8] \\
\textsc{Lucid} feedback + few-shot ($k=5$) & 87.8\% & [85.6, 89.7] \\
\bottomrule
\end{tabular}%
}
\caption{Few-shot baselines on ProcessBench. Self-correction outperforms few-shot prompting.}
\label{tab:fewshot-comparison}
\end{table}